\RequirePackage{fix-cm}
\documentclass[twocolumn]{svjour3}          
\smartqed  

\usepackage{graphicx}

\usepackage{url}
\usepackage{xcolor}
\usepackage{hyperref}
\hypersetup{
     colorlinks=true,
     linkcolor=orange,
     filecolor=orange,
     citecolor=orange,      
     urlcolor=orange,
     }
\usepackage{color, colortbl}
\usepackage{amsfonts}
\usepackage{amsmath}
\usepackage{appendix}
\usepackage{algorithm, setspace}
\usepackage{algpseudocode}
\usepackage{csquotes}
\usepackage{amsfonts}
\usepackage{booktabs}
\usepackage{siunitx}
\usepackage{breqn}
\usepackage[normalem]{ulem}
\usepackage{tikz}
\usetikzlibrary{tikzmark}

\newcommand{\p}[1]{\medskip \noindent \textbf{{#1}.}}
\newcommand{\eq}[1]{Equation~(\ref{eq:#1})}
\newcommand{\fig}[1]{Figure~\ref{fig:#1}}

\journalname{Autonomous Robots}

\DeclareUnicodeCharacter{2212}{-}

\begin{document}

\title{
Learning Latent Representations to Co-Adapt to Humans}


    \author{Sagar Parekh \and Dylan P. Losey}

\authorrunning{Short form of author list} 

\institute{S. Parekh \at
              Mechanical Engineering Department, Virginia Tech \\
              \email{sagarp@vt.edu}           
           \and
           D. Losey \at
            Mechanical Engineering Department, Virginita Tech \\
            \email{losey@vt.edu}
}

\maketitle

\begin{abstract}

When robots interact with humans in homes, roads, or factories the human's behavior often changes in response to the robot. Non-stationary humans are challenging for robot learners: actions the robot has learned to coordinate with the original human may fail after the human adapts to the robot. In this paper we introduce an algorithmic formalism that enables robots (i.e., ego agents) to \textit{co-adapt} alongside dynamic humans (i.e., other agents) using only the robot's low-level states, actions, and rewards. A core challenge is that humans not only react to the robot's behavior, but the way in which humans react inevitably changes both over time and between users. To deal with this challenge, our insight is that --- instead of building an exact model of the human --- robots can learn and reason over \textit{high-level representations} of the human's policy and policy dynamics. Applying this insight we develop RILI: Robustly Influencing Latent Intent. RILI first embeds low-level robot observations into predictions of the human's latent strategy and strategy dynamics. Next, RILI harnesses these predictions to select actions that influence the adaptive human towards advantageous, high reward behaviors over repeated interactions. We demonstrate that --- given RILI's measured performance with users sampled from an underlying distribution --- we can probabilistically bound RILI's expected performance across new humans sampled from the same distribution. Our simulated experiments compare RILI to state-of-the-art representation and reinforcement learning baselines, and show that RILI better learns to coordinate with imperfect, noisy, and time-varying agents. Finally, we conduct two user studies where RILI co-adapts alongside actual humans in a game of tag and a tower-building task.  See videos of our user studies here: \url{https://youtu.be/WYGO5amDXbQ}

\end{abstract}


\section{Introduction}

Small and mid-sized manufactures (SMMs) make up almost $99\%$ of all American manufacturing companies, and more than $75\%$ of SMMs have fewer than $20$ employees \cite{smallbusiness}. But despite their prevalence, SMMs are the least likely manufacturers to use robots \cite{sanneman2021state}. To become successful within small and mid-sized manufacturers
robots must \textit{learn} new tasks while \textit{coordinating} with human co-workers \cite{perzylo2019smerobotics,gualtieri2020opportunities,arents2022smart,matheson2019human}. 

\begin{figure}
    \centering
    \includegraphics[width=1.0\columnwidth]{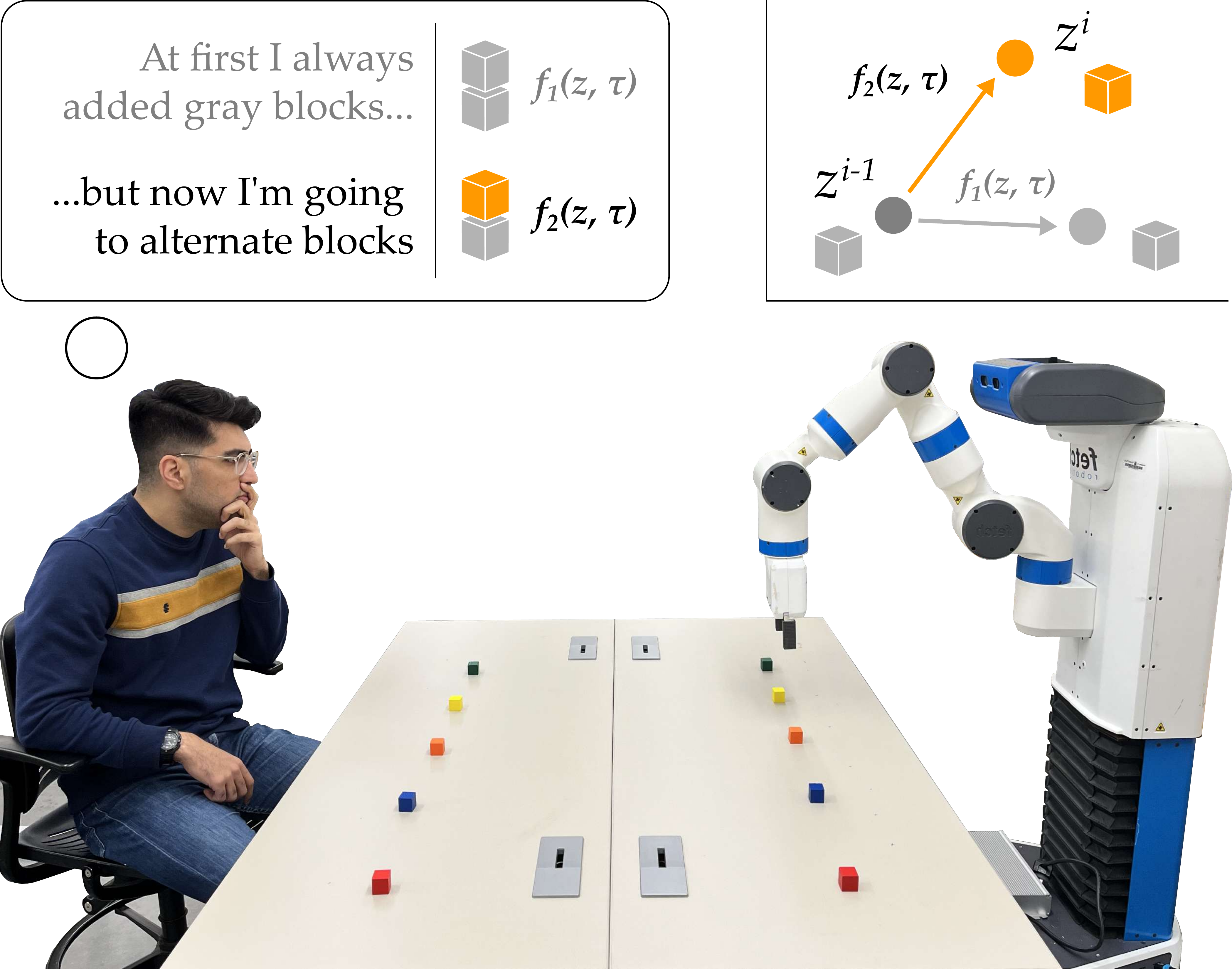}
    \caption{Human and robot interact to assemble towers. The robot's objective is to build the same tower as the human. As the robot adapts to the human, the human also adapts to the robot and changes their behavior. Different users react to the robot in different ways: even the same human will inevitably change how they respond to the robot over time. Our approach enables robots to co-adapt alongside non-stationary humans by learning a high-level representation of the human's policy (strategy $z$) and maintaining multiple models for how the user's strategy will change over time (dynamics $f_p$).}
    \label{fig:front}
\end{figure}

Consider a robot arm that is learning to build towers with a human partner (see \fig{front}). During each interaction the human and robot both add one block to their respective towers, and the robot is rewarded if its tower matches the human's. The robot's learning would be straightforward if the human always built the exact same tower regardless of what the robot did. In practice, however, humans \textit{adapt} to the robot's behavior \cite{nikolaidis2017human,goodrich2008human,ikemoto2012physical}. Perhaps a competitive human sees the robot is almost always picking green blocks, and so the human switches their behavior to now add orange blocks to the tower. Changes in human behavior present a challenge to robot learners: these shifts alter the robot's learning environment, so that robot actions which originally coordinated with the human (e.g., adding green blocks) are no longer effective \cite{hernandez2017survey}. To make matters worse, different humans adapt to the same robot behavior in different ways. While a competitive human adapts by picking blocks that the robot does not expect, a collaborative human may help the robot by selecting the same blocks that the robot chooses frequently.

\begin{figure*}
    \centering
    \includegraphics[width=2\columnwidth]{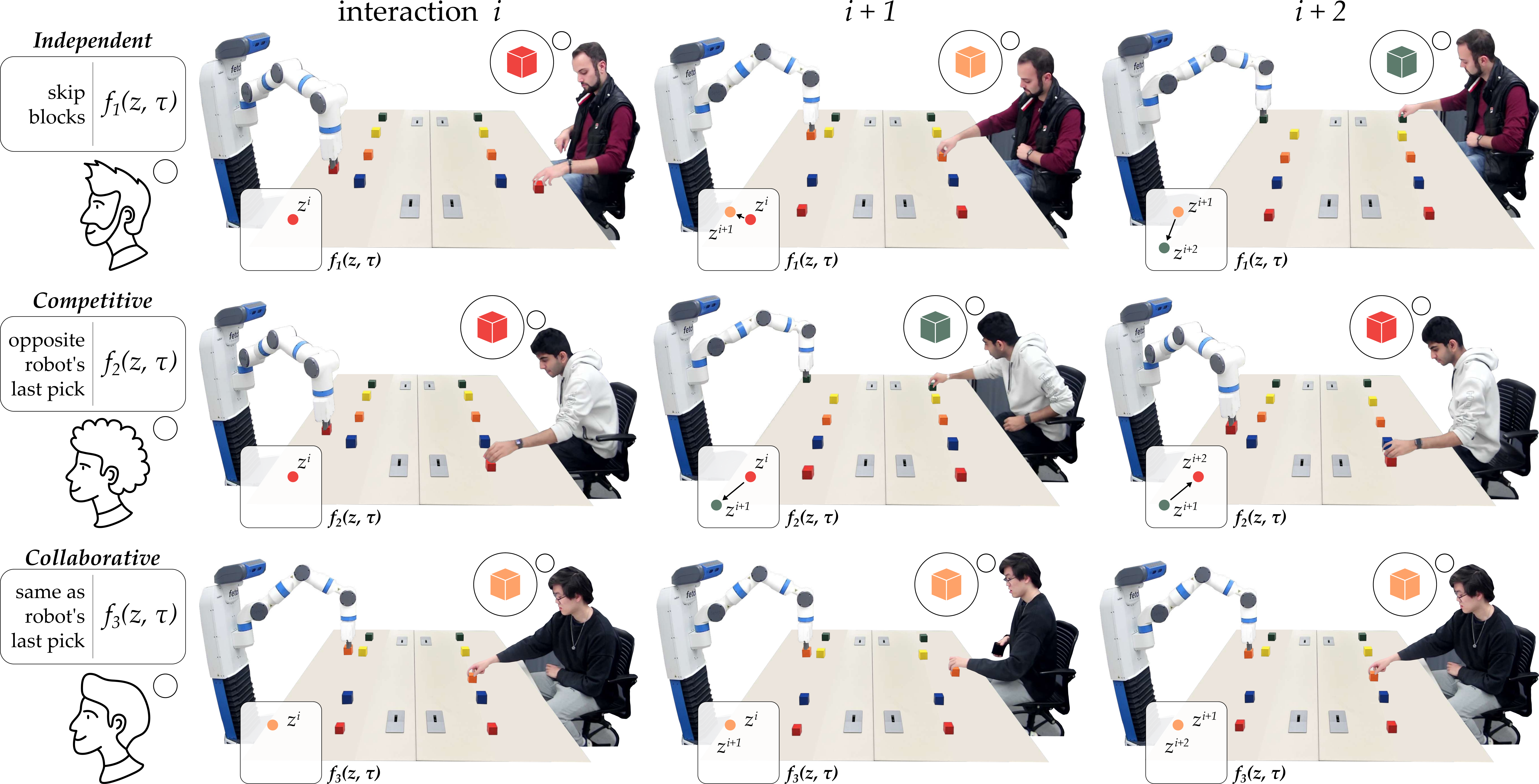}
    \caption{Robot learning to select blocks that match the human. The robot encounters different humans who react to the robot in different ways: some users ignore the robot's behavior (\textit{Independent}, Top), while others adapt to choose blocks away from the robot (\textit{Competitive}, Middle) or match the robot's last choice (\textit{Collaborative}, Bottom). Co-adaptive robots must be able to learn alongside each type of human (i.e., each different dynamics). This is particularly challenging when a single human changes dynamics (e.g., switches from collaborative to competitive) or when a new user comes along with their own personalized rules for reacting to the robot. Under our proposed approach the robot learns latent representations of the other agent's strategy $z$ and strategy dynamics $f_p$ so that it can predict which block the human will choose next and take actions accordingly. The images above are taken from our second user study (Section~\ref{sec:user_study}) where the RILI robot learns to co-adapt with each user.}
    \label{fig:front_2}
\end{figure*}

This paper explores settings where one ego agent (e.g., a robot) is learning and interacting with one other agent (e.g., a human). Specifically, we introduce an algorithmic formalism that enables the ego agent to seamlessly \textit{co-adapt} alongside another agent. Prior research on multi-agent learning and human-robot interaction makes restrictive assumptions about this other agent. For example, existing methods assume that the other agent always reacts to the ego agent in the same way \cite{xie2020learning,wang2021influencing,carroll2019utility,foerster2018learning}, the other agent communicates its intent or is trained together with the ego agent \cite{ndousse2021emergent,woodward2020learning,foerster2018counterfactual,cao2018emergent}, or the other agent's adaptation has a pre-defined structure \cite{bestick2016implicitly,li2021influencing,nikolaidis2017game,sadigh2016planning,bandyopadhyay2013intention,losey2020learning,lu2022model}. By contrast, we recognize that humans are independent, partially observable agents, and that the human's personalized response to the ego agent will change over repeated interactions.

Humans seamlessly co-adapt to other humans on a daily {basis} (e.g., imagine breaking and accelerating to maintain distance from another car). When humans learn to coordinate with another human they do not build exact models of the other agent's policy \cite{baker2011bayesian,rubinstein1998modeling,von2017minds}. Return to our tower building example: humans do not reason over every fine-grained motion of the other agent. Instead, a human worker might predict the high-level intent that shapes the other agent's actions (e.g., the human predicts the other agent will add an orange block next interaction). Accordingly, our insight is that:
\begin{center}
    \emph{Ego agents can co-adapt by maintaining} high-level representations \emph{of both the other agent's policy and the change in policy between interactions.}
\end{center}
We leverage this insight to propose our algorithm \textbf{RILI}, \textbf{R}obustly \textbf{I}nfluencing \textbf{L}atent \textbf{I}ntent. Under RILI robots learn to embed their low-level observations into latent strategies and dynamics. Here a \textit{strategy} captures how the other agent will behave during the current interaction (e.g., which block the human will choose), and \textit{dynamics} express the underlying rules the other agent is using to change their strategy (e.g., if the robot adds an orange block now, then the human will also choose an orange block during the next interaction). By reasoning over strategies and dynamics we enable ego agents to predict which behaviors will coordinate with the other agent. This prediction is \textit{robust} to noisy and imperfect humans who adapt and change their underlying dynamics over repeated interactions (see~\fig{front_2}).

Developing a high-level representation of the other agent forms the first half of our proposed approach. Next, RILI harnesses this robust predictive model to learn to \textit{influence} the other agent. Remember that our ultimate goal is robots that co-adapt alongside humans and successfully complete multi-agent tasks. We therefore train the ego agent online --- as it interacts with the other agent --- to learn a policy that maximizes its cumulative reward. Because we have equipped the robot with our predictive model, the robot can anticipate how the human will react to its choices; put another way, the robot optimizes for behaviors \textit{now} that will guide the human towards advantageous strategies \textit{in the future}. Returning to \fig{front}, the robot receives higher rewards for adding blocks that are easy for the robot to reach. Robots that apply RILI autonomously learn actions that cause the adaptive human to reach for those blocks, thereby increasing the robot's reward and improving task coordination.

Overall, we make the following contributions\footnote{Parts of this work have been published at the International
Conference on Intelligent Robots and Systems \cite{parekh2022rili}.}:

\p{Learning to Coordinate} In \textbf{Section~\ref{sec:rili}} we introduce our RILI algorithm for settings where one ego agent repeatedly interacts with an adaptive agent, and the ego agent can only observe its own low-level states, actions, and rewards. RILI combines representation and reinforcement learning to embed low-level observations into high-level predictions of the other agent's strategy and dynamics, and then reasons over these predictions when selecting robot actions. The resulting approach learns online while it is working with the other agent.

\p{Deriving Co-Adaptation Bounds} We assert that RILI can influence other agents who change their underlying dynamics. For example, perhaps a new human comes along and starts to interact with the robot, or the existing user becomes more competitive over time. Let the other agent's dynamics be sampled from some distribution $\mathcal{P}$. Given the robot's measured reward with $N$ dynamics sampled from $\mathcal{P}$, in \textbf{Section~\ref{sec:generalization-bound}} use PAC-Bayes theory to derive a probabilistic lower bound on the robot's expected reward across the unknown distribution of other agents. We then support this theoretical bound through simulated experiments.

\p{Comparing RILI to Baselines} To compare our proposed approach to the state-of-the-art we perform extensive simulations within the the environments established by prior work. Our simulations in \textbf{Section~\ref{sec:simulations}} reveal that RILI can learn to co-adapt with an arbitrary number of other agent dynamics, including other agents who react to every robot behavior, other agents who only adapt to some robot behaviors, and other agents who ignore the robot's behavior altogether. We also show that the robot can remember old partners even after training with new partners (e.g., the robot can still coordinate with previous human users), and that the RILI approach more rapidly adapts to unexpected, out-of-distribution dynamics than the baselines.

\p{{Co-Adapting with} In-Person Users} Finally, in \textbf{Section~\ref{sec:user_study}} we put RILI to the test across two experiments with actual humans. In our first user study participants play a virtual game of tag with the RILI agent. Each user chooses their own strategy for avoiding the robot, and RILI learns from scratch how to coordinate with and catch the human participants. Our second study focuses on the tower-building environment from \fig{front}. Here the robot arm is pre-trained offline with a pool of simulated agents, and the robot must adapt online to the actual human user during a total of $30$ interactions. The results of both studies suggest that RILI leads to higher rewards than state-of-the-art baselines.

\section{Related Work}

In our approach robots learn to influence humans that change and adapt over time. The robot can only observe its own states, actions, and rewards, and we do not assume any pre-defined model of the human. Instead, the robot must learn to embed its low-level observations to a high-level representation of the human's policy and policy dynamics. We emphasize that training alongside another robot or simulated human is not sufficient: because each human may respond to the robot in different ways, we need a robust robot that personalizes its behavior to the agent it is currently working with.

\p{Learning alongside Adaptive Agents} Related work on multi-agent reinforcement learning (MARL) explores settings where an ego agent trains alongside other learning agents \cite{wong2022deep}. Because the other agents change their behaviors as they learn, the ego agent's environment is non-stationary: robot actions that initially lead to high reward may suddenly lead to low rewards when the other agents adapt. To deal with this, some MARL research leverages a \textit{centralized} learning procedure where each agent has access to the same models or observations (e.g., agents are trained using a centralized critic) \cite{foerster2018counterfactual,lowe2017multi,NEURIPS2020_7967cc8e}. Other research learns explicit communication protocols to share information between agents (e.g., agents send messages to one another) \cite{cao2018emergent,NIPS2016_c7635bfd,singh2018learning}. Neither of these methods apply to our human-robot setting where (a) the agents are decentralized and (b) we do not assume that the human and robot have a mutually understood channel for communication.

\p{Modeling Humans and Other Agents}
More relevant here is prior work from MARL and human-robot interaction that attempts to model the other agent. Within these methods the ego agent observes the other agent's behavior during \textit{previous} interactions, and learns a model to predict the other agent's \textit{future} actions. Often these models rely on some underlying structure. For instance, the robot may assume that it knows the other agent's learning rule \cite{foerster2018learning,lu2022model}, that the other agent acts as a leader, follower, or expert \cite{losey2020learning,li2021influencing,ndousse2021emergent}, or that the other agent has a fixed, unchanging policy \cite{carroll2019utility}. Research like \cite{jeon2020reward,raileanu2018modeling,sadigh2016planning} assumes that the human's behavior noisily optimizes their reward, and this reward is a function of known features (e.g., the human's goal position). Going one step further, robots can recognize that the other agent is also learning from them, resulting in recursive reasoning \cite{baker2011bayesian,von2017minds} and co-adaptation \cite{nikolaidis2017game}. {In the domain of co-adaptation previous works have used examples of expert coordination to learn a low-rank subspace over human strategies \cite{R1} or cluster human actions to learn common reward functions \cite{R2}.} Similar to prior works we will learn a predictive model of the other agent over repeated interactions. We do not assume access to the other agent's learning model, roles, reward, features, or examples of expert coordination; instead, we hypothesize that the other agent's policy is \textit{parameterized} by a high-level strategy. By learning this representation --- and predicting how the human's strategy changes in response to the robot --- we enable the ego agent to account for its non-stationary environment.

\p{Influencing Humans} Because the other agent adapts to the robot's actions, we can leverage the robot's behavior to intentionally influence the other agent. This influential behavior is not explicitly programmed as part of the robot's objective \cite{jaques2019social,lu2022model} --- under RILI influence emerges \textit{naturally} as the robot tries to guide the human towards advantageous strategies and maximize its own reward. Previous works have explored how robots can influence humans during a single interaction. In \cite{sadigh2016planning,tian2022safety,hu2022active,bestick2016implicitly,sagheb2022towards} human-robot interaction is formulated as a two-player game: the robot infers the human's reward, and then plans while accounting for the human's optimal response (e.g., an autonomous car changing lanes to slow a human driver). Most relevant here are methods like LILI \cite{xie2020learning} and SILI \cite{wang2021influencing} that learn to influence another agent over \textit{repeated} interactions. Both \cite{xie2020learning} and \cite{wang2021influencing} assume that the other agent maintains a fixed set of rules for reacting to the ego agent. Return to our motivating example: for a given robot choice (e.g., adding a green block) LILI and SILI assume that every human reacts in the exact same way (e.g., choosing a blue block). Accordingly, while \cite{xie2020learning,wang2021influencing} are effective in robot-robot experiments, we will demonstrate that these approaches fail to influence \textit{actual humans}.

\p{Robust Human Prediction} 
For our proposed approach to coordinate with actual humans it must be robust to noisy and imperfect agents who change their dynamics. Today's robots take a step towards robust interaction by maintaining a probabilistic model over the human's possible actions, and identifying risk-aware robot behaviors that achieve high rewards even for unexpected actions \cite{huang2022seamless,bajcsy2020robust,li2021provably,nishimura2021rat}. When a human model is not available (or when humans deviate from pre-defined models), robots can achieve robust performance by observing and interacting with a distribution of real or simulated agents \cite{strouse2021collaborating,jaderberg2019human,carroll2019utility,woodward2020learning}. RILI draws inspiration from both of these approaches: we learn a posterior distribution over the human's strategy through repeated interactions with multiple other agents. To analyse the robustness of our resulting algorithm we turn to Probably Approximately Correct or PAC-Bayes bounds \cite{mcallester1999some} that have been used for studying generalization in deep learning \cite{neyshabur2017exploring,cherief2022pac}. PAC-Bayes theory has also been used to optimize for robust policies in novel environments \cite{majumdar2021pac,pmlr-v87-majumdar18a}. Here we similarly leverage PAC-Bayes theory to obtain a probabilistic lower bound on performance across new and unseen human dynamics.

\section{Problem Statement}
\label{sec:problem}

We consider two-agent settings with an \textit{ego agent} and some \textit{other agent}. We control the ego agent, but the other agent is fully autonomous; for example, the ego agent could be a robot while the other agent is a human. Our approach applies to scenarios where the other agent is collaborative (i.e., a partner) or competitive (i.e., an opponent). In this paper we use ego agent or robot to refer to the agent that we control, and other agent or human to refer to the agent we are interacting with.

The ego agent \textit{repeatedly interacts} with the other agent. Recall our motivating example where an industrial robot arm is building towers with a human worker: the human and robot will need to manufacture multiple towers over hours, days, or weeks of interaction. Each time that the human and robot interact we assume that the human has some high-level \textit{strategy} $z \in \mathcal{Z}$ that they use to make low-level decisions. For instance, the human may want to build a tower with the purple block above the orange block. Let $i$ be the current interaction: during interaction $i$, the human uses strategy $z^i$ to reach for, pick up, and move the blocks. {The human's high-level strategy changes between interactions according to their underlying \textit{dynamics}.} Perhaps the human noticed that the robot expects the purple block to be below the orange block, and so during interaction $i+1$ the human changes their high-level strategy $z^{i+1}$ to also place the orange block above the purple block. 

Importantly, not all other agents follow the same dynamics to update $z$. Returning to our example, some humans may change their high-level strategy to build a tower that matches the robot's behavior; other humans may ignore the robot entirely and build the towers that they prefer. Even a single human's dynamics will inevitably shift over time --- causing the human to respond to the same robot behavior in different ways. In this section we therefore formalize two-agent interactions where the other agent has a high-level strategy $z$ as well as changing dynamics for updating that strategy. We emphasize that the ego agent can never directly observe the high-level strategy $z$, and the ego agent does not know what dynamics the human is using to update $z$ between interactions.

\p{Latent Strategy} We start by formulating a single interaction. Every interaction lasts a total of $H$ timesteps; at each timestep the robot observes its state $s \in \mathcal{S}$ and takes action $a \in \mathcal{A}$. The robot does not know the latent strategy of the other agent $z \in \mathcal{Z}$. However, this high-level strategy affects how the other agent behaves, and this in turn may alter what the robot observes. More specifically, both the ego agent's transition function $\mathcal{T}(s' \mid s, a, z^i)$ and reward function $R(s, z^i)$ depend on the current latent strategy $z^i$. Consider the running example of a robot trying to assemble towers that match the human; the tower that the human chooses to build determines how much reward the robot receives. By combining these parts we express a single interaction as a Hidden Parameter Markov Decision Process (HiP-MDP) using the tuple $\mathcal{M} = \langle \mathcal{S}, \mathcal{A}, \mathcal{Z}, \mathcal{T}, R, H\rangle$ where $z \in \mathcal{Z}$ is the hidden parameter. During the $i$-th interaction the ego agent follows the state-action \textit{trajectory} $\xi^i = \{(s_1^i, a_1^i), \ldots, (s_H^i, a_H^i)\}$, and the ego agent observes this trajectory and its rewards $r$ at every timestep. Let $\tau^i = \{(s_1^i, a_1^i, r_1^i), \ldots, (s_H^i, a_H^i, r_H^i)\}$ be the robot's \textit{experience}. We emphasize that trajectory $\xi^i$ and experience $\tau^i$ contain only low-level information on the states, actions, and rewards of the ego agent.

\p{Latent Dynamics} Within a single interaction the other agent maintains a constant latent strategy. But between interactions this strategy changes according to the human's latent dynamics:
\begin{equation} \label{eq:P1}
    z^{i+1} = f_p(z^i, \tau^i)
\end{equation}
Imagine a person building towers with the robot in \fig{front}. The person updates their choice of $z$ based on some personalized set of rules: they may choose to build the same tower every time, cycle through different choices of towers, or even change the tower they build in response to how the robot behaves. Each of these cases corresponds to a different dynamics function $f$. We capture these differences using subscript $p$, so that the $p$-th other agent has latent dynamics $f_p$.

We next recognize that --- not only are there many possible latent dynamics --- but these dynamics will inevitably change over repeated interactions. This could be because the robot is now interacting with a new agent (i.e., the robot starts working with a different human) or because the other agent changes (i.e., the same human modifies how they react to the robot). In either case, the robot interacts with multiple dynamics $f_p$, where $p \sim \mathcal{P}$ is sampled from a distribution over other agents. From the ego agent's perspective the dynamics shift randomly: for the first $m$ interactions the robot may interact with dynamics $f_{p_1}$, then the next $n$ interactions the robot may interact with dynamics $f_{p_2}$. The ego agent cannot observe the latent dynamics and does not know when the latent dynamics change. Similarly, the ego agent does not know the distribution $\mathcal{P}$ from which these different dynamics are sampled. \eq{P1} and the listed assumptions describe scenarios where the robot must learn to interact with different humans (which could be competitive, collaborative, or indifferent), and the robot does not know how these other agents will behave \textit{a priori}.

\p{Repeated Interaction} We have separately discussed the latent strategy \textit{within} an interaction and how that latent strategy changes \textit{between} interactions. Putting these together we reach our problem formulation. Over {$n$} repeated interactions the ego agent encounters a sequence of {$n$ HiP-MDPs: $(\mathcal{M}_1, \ldots, \mathcal{M}_n)$} where the other agent plays strategy $z^i$ throughout $\mathcal{M}_i$. The other agent uses latent dynamics $f_p$ to update $z$ between interactions, and the ego agent interacts with a total of {$N$} different latent dynamics, where {$N \leq n$}. {To clarify: $n$ indexes the number of interactions and $N$ is the number of different latent dynamics the ego agent has so far encountered.} In practice, these $N$ dynamics could correspond to interacting with one agent that changes their dynamics {$N$} times, {$N$} different humans who each have their own approach for interacting with the robot, or some combination of the above. 

The robot's total reward is the sum of the rewards across all interactions, and the robot's objective is to maximize its total reward. Maximizing reward often requires that the robot \textit{influence} the other agent so that they choose latent strategies with which the robot can seamlessly coordinate. In our motivating example there are some towers that are easier for the robot to build (e.g., the robot receives more reward for specific towers). The robot can therefore increase long-term reward by guiding the human towards strategies that correspond to these towers. Efficient ego agents will therefore i) identify the other agent's current dynamics $f_p$ and ii) exploit those dynamics to influence the other agent towards advantageous latent strategies $z$.

\section{Robustly Influencing Latent Intent (RILI)}
\label{sec:rili}

\begin{figure*}
    \centering
    \includegraphics[width=2\columnwidth]{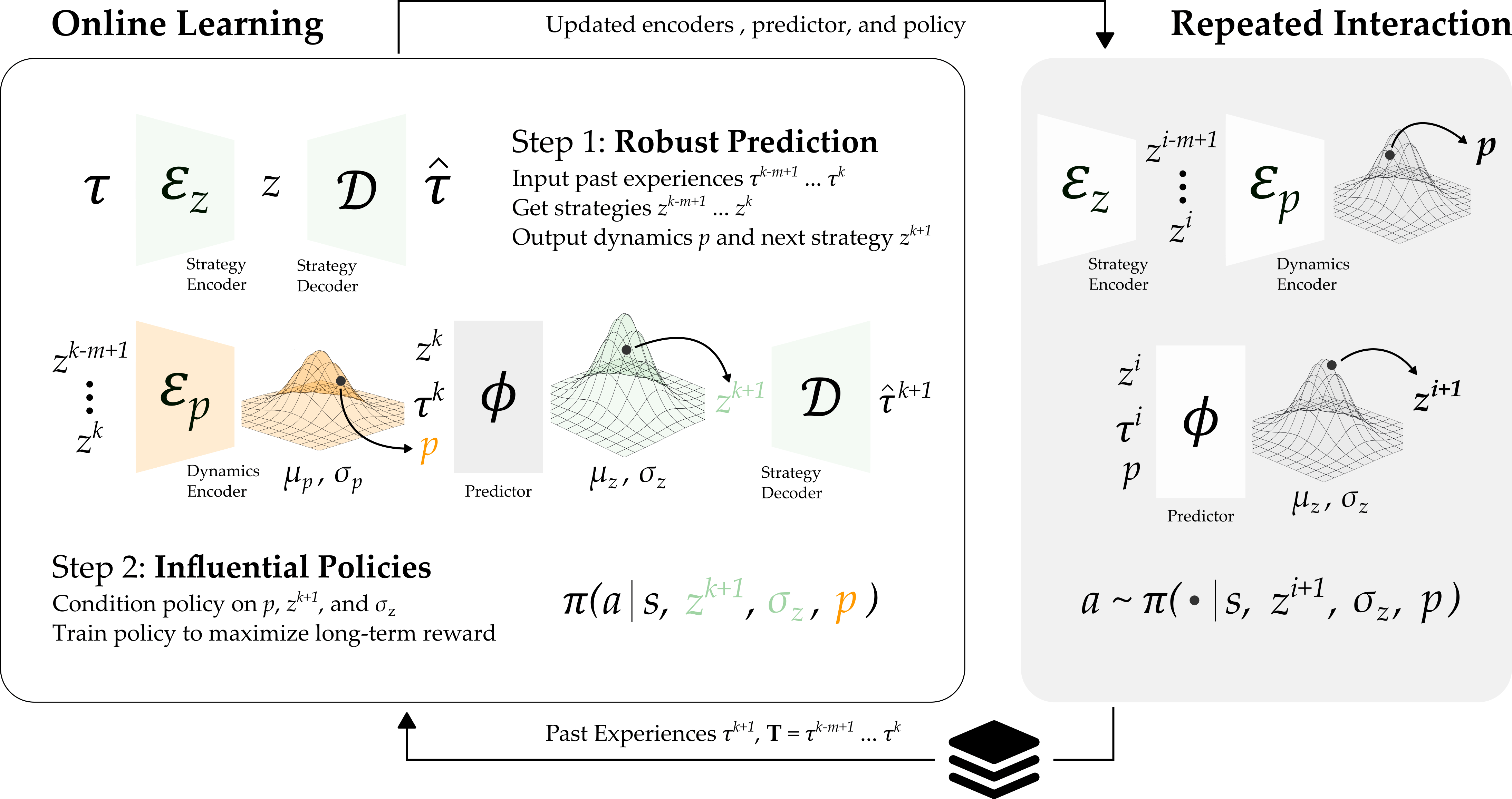}
    \caption{Robustly Influencing Latent Intent. (Left) RILI learns online while interacting with another agent. Given a sequence of past experiences, the robot learns to embed each interaction into a strategy $z$ and the sequence of strategies into dynamics $p$. The robot then predicts how an agent with dynamics $p$ selected their next strategy $z^{k+1}$ given their current strategy $z^k$ and interaction $\tau^k$. The encoders and the predictor are trained using a strategy decoder that reconstructs $\tau$. We finally condition the robot's policy on our high-level representation of the other agent, and leverage model-free, off-policy reinforcement learning to maximize long-term reward. (Right) At interaction $i$ the updated models are leveraged to predict the strategy and dynamics of the other agent for interaction $i+1$. The robot co-adapts to the human by taking actions based on the robot's prediction. Note that the dynamics encoder and predictor output a mean and standard deviation over the latent representation.}
    \label{fig:method}
\end{figure*}

In this section we present \textbf{R}obustly \textbf{I}nfluencing \textbf{L}atent \textbf{I}ntent (\textbf{RILI}), our proposed approach for co-adapting alongside another agent with changing latent dynamics. RILI breaks down into two components. First, in Section~\ref{sec:M1} the ego agent learns  to \textit{predict} how the other agent will respond to the robot's behaviors. Given the robot's low-level states, actions, and rewards during previous interactions, can the robot anticipate the human's high-level strategy for the current interaction? Making accurate predictions is challenging because the dynamics the other agent uses to choose its strategy will inevitably shift over repeated interaction, and the ego agent cannot observe either dynamics or high-level strategies. Second, in Section~\ref{sec:M2} the ego agent leverages these predictions to \textit{influence} the human's strategy over repeated interactions. Given that the robot has a model of how the human will react to its actions, which actions should the robot select to exploit the human's latent dynamics and maximize its long-term reward? Overall, RILI combines representation and reinforcement learning to continually adapt to changing partners: see the method outline in \fig{method}.

\subsection{Robust Prediction}\label{sec:M1}

Our first challenge is predicting how the human will react to the robot's actions, i.e., anticipating the human's next latent strategy. Consider our running example: if the robot can accurately predict which block a human will choose next, the robot can seamlessly coordinate with that human. We know that the other agent's next latent strategy $z^{i+1}$ is selected according to \eq{P1}. Here we therefore enable the robot to learn a model of \eq{P1} across all dynamics $p$ that the robot has encountered so far.

\p{Inferring Strategies} To start, we recognize that the human's next latent strategy $z^{i+1}$ is a function of their current strategy $z^i$ and the interaction experience $\tau^i$. The ego agent directly observes the states, actions, and rewards $\tau^i = \{(s_1^i, a_1^i, r_1^i), \ldots, (s_H^i, a_H^i, r_H^i)\}$, but the other agent's strategy $z^i$ is hidden from the robot. Recall that the transition and reward functions during interaction $i$ depend on latent strategy $z^i$. As a result, the robot can leverage the states, actions, and rewards in $\tau^i$ to reconstruct $z^i$; e.g., based on the rewards the robot receives, the robot can determine which block the human picked up. We introduce the \textit{strategy encoder}:
\begin{equation}\label{eq:z_i}
    z^i = \mathcal{E}_z(\tau^i), \quad z \in \mathbb{R}^d
\end{equation}
where $\mathcal{E}_z$ maps the $i$-th interaction to a representation of the agent's strategy $z^i$. Because the actual strategies of the other agent are never observed, we cannot train this encoder using ground-truth labels.

Instead, we next introduce a \textit{strategy decoder}. This decoder attempts to reconstruct the robot's rewards when the robot executes trajectory $\xi$ and the other agent has latent strategy $z \in \mathbb{R}^d$:
\begin{equation}
    [\hat{r}_1^i, \ldots \hat{r}_H^i]^T = \mathcal{D}(\xi^i, z^i)
\end{equation}
where $\xi^i = \{(s_1^i, a_1^i), \ldots, (s_H^i, a_H^i)\}$ includes the ego agent's states and actions during interaction $i$, and $\hat{r}$ are rewards predicted by the decoder. Pairing the strategy encoder and decoder, we reach the loss function:
\begin{equation}\label{eq:encoder_z_loss}
    \mathcal{L}_z = \left \| \begin{bmatrix}r_1^i \\ \vdots \\ r_H^i \end{bmatrix} - \mathcal{D} \Big(\xi^i, \mathcal{E}_{z}(\tau^{i})\Big) \right \|
\end{equation}
The loss $\mathcal{L}_z$ is minimized when the decoder accurately reconstructs rewards. Intuitively, this means the encoder $\mathcal{E}_z$ must output a latent strategy $z$ that captures enough information about the other agent such that --- given $z$ --- the robot can correctly score its own behavior $\xi$. For instance, if the reward function is based on the distance between $\xi$ and the other agent (e.g., the distance between the robot and human's blocks), then $z$ should implicitly represent the other agent's position.

\p{Inferring Dynamics} Let us return to \eq{P1}. We have a method for inferring the current strategy; but just knowing $z$ and $\tau$ is not sufficient to accurately predict $z^{i+1}$. We need to know the other agent's dynamics, and these dynamics will inevitably change over time --- either because the robot encounters a new agent, or because the same agent starts reacting in a different way. We capture these unique dynamics in \eq{P1} using $p \sim \mathcal{P}$, i.e., dynamics $f_{p_i}$ are different from dynamics $f_{p_j}$. In practice, the ego agent does not know when the other agent will change their dynamics and shift how they respond to the robot. Instead of learning separate models for each $p \sim \mathcal{P}$, we therefore capture the other agent's dynamics through a single model:
\begin{equation}\label{eq:F}
    z^{i+1} = \phi(p, z^i, \tau^i), \quad p \in \mathbb{R}^d
\end{equation}
where $p \in \mathbb{R}^d$ is the robot's latent representation of the other agent's current dynamics (i.e., the other agent's type), and $\phi$ uses this dynamics representation to predict $z^{i+1}$. Given $z$ and $\tau$, different choices of $p$ result in different predictions of $z^{i+1}$. Returning to our motivating example: for one latent dynamics $p$ the robot may predict that the human will pick up the red block during the next interaction, while for another $p$ the robot predicts the human will reach the blue block.

To infer the latent dynamics $p$ we look back at the other agent's behavior over the last $m$ interactions. Inverting \eq{P1}, we can solve for $p$ based on the past sequence of strategies $z$ and experiences $\tau$. We approximate this using a \textit{dynamics encoder}:
\begin{equation} \label{eq:de}
    p = \mathcal{E}_p(h^i), \quad h^i = \{z^i, z^{i-1}, \cdots, z^{i-m+1}\}
\end{equation}
where $h$ is the history of $m$ strategies and $\mathcal{E}_p$ embeds this history to a representation of the agent's dynamics $p$. Note that this approach i) assumes the other agent's dynamics remain constant across $h$ and ii) does not include $\tau^{i}\, \ldots, \tau^{i-m+1}$ within $h$. We leave out $\tau$ because $z$ is already an embedding of $\tau$ from \eq{z_i}, and because this functional approximation works well across our simulations and experiments.

Now that we have developed an encoder to infer $p$, we can use $\phi$ to predict what latent strategy the current human will follow during the next interaction. Of course, the robot cannot observe the actual dynamics the other agent uses to select their high-level strategies. We therefore leverage the robot's low-level observations and our strategy decoder $\mathcal{D}$ to learn the dynamics encoder $\mathcal{E}_p$ in \eq{de} and the overall dynamics $\phi$ in \eq{F}. Given $\tau^{i}\, \ldots, \tau^{i-m+1}$, we first apply $\mathcal{E}_z$ to recover $z^{i}\, \ldots, z^{i-m+1}$. We then use this sequence to predict $z^{i+1}$, before finally decoding $z^{i+1}$ to estimate the ego agent's rewards during interaction $i+1$:
\begin{equation}\label{eq:F_loss}
    \mathcal{L}_\phi = \left \| \begin{bmatrix}r_1^{i+1} \\ \vdots \\ r_H^{i+1}\end{bmatrix} - \mathcal{D} \bigg(\xi^{i+1}, \phi\Big(\tau^{i}, \mathcal{E}_p(h^i), \mathcal{E}_z(\tau^i)\Big)\bigg) \right \|
\end{equation}
In our running example the robot is assembling towers with a human, and the robot's reward is the distance between the block it selected and the block the human selected. To accurately estimate this reward the robot must correctly anticipate which block the human will choose; in \eq{F_loss}, this means $z^{i+1}$ must learn to correctly capture the human's \textit{next} block.

\p{Representation Learning} Now that we have introduced the individual components of our prediction framework, we will discuss how to train the robot to predict the other agent's strategy. Here training involves learning the weights of the strategy encoder $\mathcal{E}_z$, dynamics encoder $\mathcal{E}_p$, predictor $\phi$, and strategy decoder $\mathcal{D}$. We structure the dynamics encoder and the predictor as conditional variational autoencoders. Specifically, $\mathcal{E}_p$ outputs the mean $\mu_p$ and standard deviation $\sigma_p$ over the latent space $p$, while $\phi$ outputs the mean $\mu_z$ and standard deviation $\sigma_z$ over the latent strategy space $z^{i+1}$. Define the overall mean and standard deviation as $\mu = (\mu_p, \mu_z)$ and $\sigma = (\sigma_p, \sigma_z)$. In practice, higher values of $\sigma$ indicate that the robot is uncertain about its prediction, while lower values of $\sigma$ suggest that the robot is confident about $p$ and $z^{t+1}$. Our overall loss function for robust strategy prediction sums the reconstruction losses \eq{encoder_z_loss} and \eq{F_loss} with a regularization term that enforces a $\mathcal{N}(0, 1)$ Gaussian prior over the latent space:
\begin{equation}\label{eq:total_loss}
    \mathcal{L} = \sum_{T \in \mathcal{B}} \Bigg( \mathcal{L}_z + \mathcal{L}_\phi + \underbrace{KL\Big( \mathcal{N}(\mu, \sigma) \mid\mid \mathcal{N}(0, 1) \Big)}_{\text{regularizer}}\Bigg)
\end{equation}
Here $T = (\tau^k, \ldots, \tau^{k+m})$ is a sequence of consecutive interactions and $\mathcal{B}$ is the memory buffer that contains past interactions. We emphasize that our prediction models are not just trained once; we apply the loss function in \eq{total_loss} throughout each interaction to continually improve the ego agent's ability to anticipate the other agent's response.

\subsection{Influential Policies}\label{sec:M2}

In the first half of our RILI approach we developed a representation learning structure that enables robots to predict the other agent's next strategy. The second half of our RILI approach \textit{harnesses} these predictions to influence the other agent towards strategies the ego agent can exploit. Robots that anticipate how humans will react to their behaviors can choose actions to intentionally shape the human's response. We learn to influence others without hand-coded policies or heuristics: instead, the robot uses reinforcement learning to identify high reward behaviors, and \textit{influence} becomes a natural outcome of this optimization procedure.

\p{Robot Policy} In our motivating example a robot is trying to build towers with a human. The robot needs to determine which block to reach for (i.e., which actions $a \in \mathcal{A}$ to take). To collaborate and pick up the same block as the human, the robot must first anticipate the human's latent strategy $z$ during the current interaction $i$. For instance, if the human's $z^i$ is reaching for the blue block, then the robot should also reach for blue; but if instead the human's $z^i$ is reaching for a red block, then the robot needs to take different actions. Because different human strategies require different robot responses, we assert that the robot's policy should depend on the predicted $z$. But the ego agent also needs to understand what dynamics $p$ the human will follow when reacting to the robot's behavior. For instance, if the current human updates their strategy to pick up whichever block the robot grasped during the last interaction, then the robot can leverage this knowledge of $p$ to guide the other agent. Accordingly, we learn a policy{, parameterized by weights $w$,} that is conditioned on the predicted strategy, dynamics, and the robot's uncertainty over this prediction:
\begin{equation} \label{eq:policy}
    a \sim {\pi_w}(\, \cdot \mid s, z^i, p^i, \sigma^i)
\end{equation}
Note that $z^i$, $p^i$, and $\sigma^i$ are held constant throughout the $i$-th interaction, but are updated between interactions using the models from Section~\ref{sec:M1}. Including $\sigma$ enables the ego agent to take actions that are collaborative with a distribution of strategies when the robot is unsure about its prediction (i.e., picking up the block the human has reached for most often).

\p{Reinforcement Learning} We train the robot's policy to maximize the ego agent's reward across repeated interaction. Specifically, the ego agent learns a policy with weights $\omega$ in order to maximize:
\begin{equation}\label{eq:reward}
    \max_{\omega} \sum_{i=1}^\infty  \left[ \gamma^i \mathbb{E}_{\rho^i_{\omega}} \left[ \sum_{t=1}^H r(s_t, z^i) \right] \right]
\end{equation}
Here $\gamma \in [0, 1)$ is the discount factor, $\rho^i_{\omega}$ is the distribution over trajectories $\xi^i$ under a policy with weights $\omega$, and $z^i$ is the other agent's true strategy that transitions according to dynamics \eq{P1}. Ego agents that maximize \eq{reward} will naturally influence the other agent towards advantageous strategies $z$. In our recurrent example the human's strategy is the block that they choose, and there are some blocks that are easier for the robot to reach (e.g., blocks closer to the robot). Since the ego agent receives the most reward during the $i$-th interaction if the human and robot reach for the closest block, the robot is encouraged to learn a policy that guides the human towards this $z$. Importantly, influential behavior is made possible by the robot's predictions $z$ and $p$. Because the robot anticipates how the other agent will respond to its actions, it learns to select actions \textit{now} that influence the human towards beneficial strategies \textit{in future interactions}.
\section{Lower Bounds on RILI Performance}
\label{sec:generalization-bound}

In Section~\ref{sec:rili} we introduced RILI, our approach for influencing humans whose latent dynamics change over time. If the robot trains with $N$ other agents across many interactions, we would expect the robot to coordinate efficiently with these $N$ agents. Take our motivating example of a robot learning to build towers with a human: after practicing with this specific human for many hours, days, or weeks, the robot should accurately anticipate the human's blocks and influence their choices. In practice, however, the robot will inevitably encounter \textit{new dynamics} (e.g., new humans) that respond to the robot's actions in different or unexpected ways. Given the robot's performance with $N$ latent dynamics sampled from an underlying distribution $\mathcal{P}$, how will the robot perform with other agents sampled from this same distribution? Here we answer this question by finding Probably Approximately Correct-Bayesian bounds (\textit{PAC-Bayes}) \cite{mcallester1999some}. PAC-Bayes theory has been shown to provide the tightest known generalization bounds for supervised learning problems \cite{germainPAC,langford_2003PAC,seeger2002pac}. In Section~\ref{sec:pac_prelim} we first overview an existing PAC-Bayes bound for supervised learning settings. Next, in Section~\ref{sec:pac_ours} we develop a correspondence between the RILI algorithm and these supervised learning settings, and then leverage this correspondence to extend the PAC-Bayes bound to our RILI algorithm. The result is a probabilistic lower bound on the robot's expected reward across a distribution $\mathcal{P}$ of latent dynamics. This bound depends on the number of dynamics the ego agent has encountered, the ego agent's measured performance with these seen dynamics, and the divergence between the prior and posterior of the latent space. Later, in Section~\ref{sec:simulations} we will put the bound to the test, and numerically show that the RILI agent's expected performance matches the theoretical bound.

\subsection{Preliminaries of PAC-Bayes}\label{sec:pac_prelim}

PAC-Bayes theory is used to derive PAC (Probably Approximately Correct) bounds for learning algorithms \cite{neyshabur2017exploring,mcallester1999some,germainPAC,langford_2003PAC}. In this section we present a brief overview of PAC-Bayes bounds for supervised learning settings; in the next subsection we will extend this bound to our RILI algorithm. 

\p{Supervised Learning} Consider a robot that is learning to label inputs $x \in \mathcal{X}$. Let $y \in \mathcal{Y}$ be the space of labels, and assume the robot maps inputs to predicted labels using a function parameterized by weights $\sigma$. In other words, $\hat{y} = \sigma(x)$, where $\hat{y}$ is the robot's predicted label. At the start of the task the robot has a prior $P_0(\sigma)$ over the weights. But as the robot observes new inputs $x$ --- and their true labels $y$ --- it refines its belief over $\sigma$. Let $\mathcal{P}(x)$ be the distribution over inputs, and let $\mathcal{S} = \{x_1, \ldots x_N\}$ be $N$ samples drawn from this distribution. Given these $N$ samples and their true labels $\{y_1, \ldots y_N\}$ the robot learns a posterior $P(\sigma)$ over the model weights. Ideally, the robot will learn choices of $\sigma$ that correctly label the inputs. Define $L\big(\sigma(x), y\big)$ as the loss function, where $L$ captures the error between the robot's prediction $\hat{y} = \sigma(x)$ and the true label $y$. Without loss of generality we assume that this loss function is normalized so that $0 \leq L\big(\sigma(x), y\big) \leq 1$.



\subsection{PAC-Bayes Bounds for RILI}\label{sec:pac_ours}
\begin{table*}[]
    \centering
    \begin{tabular}{lcclc}
    \hline
        \multicolumn{2}{c}{\textbf{Supervised Learning}} & & \multicolumn{2}{c}{\textbf{RILI}} \\
        \hline
        Input data & $x \in \mathcal{X}$ & $\rightarrow$ & Dynamics & $f_p \sim \mathcal{P}$ \\
        Input data distribution & $\mathcal{P}(x)$ & $\rightarrow$ & Dynamics distribution & $\mathcal{P}(f_P)$ \\
        Model weights & $\sigma$ & $\rightarrow$ & Latent variable & $\theta = (z, p)$ \\
        Loss & $L(\sigma(x), y)$ & $\rightarrow$ & Cost & $C(\theta, f_p)$ \\
        Encountered inputs & $\mathcal{S} = \{x_1, \ldots, x_N\}$ & $\rightarrow$ & Encountered dynamics & $\mathcal{S} = \{f_{p_1}, \ldots, f_{p_N}\}$ \\
        \hline
    \end{tabular}
    \caption{Correspondence between supervised learning and our proposed RILI algorithm. The supervised learning variables are defined in Section~\ref{sec:pac_prelim} and the RILI variables are discussed in Section~\ref{sec:pac_ours}. We develop this correspondence so that we can apply PAC-Bayes theory to our RILI algorithm. This results in \eq{gen_bound}, a lower bound on expected reward (upper bound on expected cost) across a distribution of dynamics $\mathcal{P}$ based on the measured rewards with $N$ encountered dynamics.}
    \label{tab:analogy-PAC}
\end{table*}

\p{Existing Theory} Within this supervised learning context, existing work \cite{maurer2004note,mcallester1999some} has derived a probabilistic upper bound on the robot's loss. For any $\delta \in (0, 1)$, with probability at least $1 - \delta$ we have that:
\begin{equation}\label{eq:pac_bound}
    L_{\mathcal{P}}(P) \leq L_{\mathcal{S}}(P) + \sqrt{ \frac{\text{KL}(P \mid\mid P_0) + \log\left( \frac{2\sqrt N}{\delta} \right)}{2N}}
\end{equation}
Here $\text{KL}(P \mid\mid P_0)$ is the Kullback–Leibler divergence between the posterior $P(\sigma)$ and the prior $P_0(\sigma)$. The measured loss $L_{\mathcal{S}}$ is the expected loss across the $N$ datapoints that the robot has already seen, and $L_{\mathcal{P}}$ is the expected loss across the entire distribution --- including inputs $x \sim \mathcal{P}(x)$ that the robot has not interacted with. Formally, these losses are defined as:
\begin{equation} \label{eq:L_S}
    L_{\mathcal{S}}(P) = \frac{1}{N} \sum_{i=1}^N \mathop{\mathbb{E}}_{\sigma \sim P} L\big(\sigma(x_i), y_i\big)
\end{equation} \vspace{-0.5em}
\begin{equation} \label{eq:L_D}
    L_{\mathcal{P}}(P) = \mathop{\mathbb{E}}_{x \sim \mathcal{P}} \mathop{\mathbb{E}}_{\sigma \sim P} L\big(\sigma(x), y\big)
\end{equation}
We note that \eq{L_D} is not actually evaluated in practice. Instead, by empirically calculating the loss $L_{\mathcal{S}}$ on the $N$ inputs we \textit{have} seen, we leverage \eq{pac_bound} to obtain an upper bound on \eq{L_D}. This enables the robot to confidently generalize its performance: using the losses the robot has measured, the robot can bound its expected performance across the entire distribution of inputs $\mathcal{P}(x)$.


Now that we have reviewed a key PAC-Bayes bound for supervised learning, we will extend the theory to our RILI algorithm. The purpose of this analysis is to quantify how RILI \textit{generalizes}: given the network's performance with $N$ different humans, how will RILI perform with new agents that have dynamics sampled from the same underlying distribution $\mathcal{P}$?

To reach our generalization result we will develop a precise analogy between the supervised setting in Section~\ref{sec:pac_prelim} and our proposed approach. Our robot interacts with another agent that has latent dynamics $f_p \sim \mathcal{P}$, where distribution $\mathcal{P}$ is not known by the robot. These dynamics are updated over repeated interactions: consider a robot that has interacted with $N$ different dynamics (e.g., $N$ different humans) such that $\mathcal{S} = \{f_{p_1}, \ldots, f_{p_N}\}$. For each human the robot records multiple sequences of experiences. Let $T_k^i = \{\tau_k^i, \ldots, \tau_k^{i-m+1}\}$ be the $i$-th \textit{sequence} of interactions with the $k$-th other agent, and let $T_k = \{T_k^1, T_k^2, \ldots\}$ contain all sequences for the $k$-th agent. Intuitively, these sequences are batches of low-level data that RILI uses to predict the other agent's strategy and select the robot's actions. 

Recall that the representation learning component of RILI (Section~\ref{sec:M1}) inputs $T_k^i = \{\tau_k^i, \ldots, \tau_k^{i-m+1}\}$ and uses the strategy encoder, dynamics encoder, and predictor to estimate the human's strategy $z$ and dynamics $p$. To capture both $z$ and $p$ we introduce a new variable $\theta = (z, p)$. More specifically, our conditional variational autoencoders in \eq{total_loss} output a Gaussian \textit{posterior} over the space of $\theta$ given input sequence $T$, such that $P(\theta \mid T) = \mathcal{N}(\mu, \sigma \mid T)$. The robot starts with a unit Gaussian prior $P_0 = \mathcal{N}(0, 1 \mid T)$ imposed by the regularization term for these variational autoencoders, and learns to map different $T$ to different means $\mu$ and standard deviations $\sigma$. Next, the reinforcement learning part of RILI (Section~\ref{sec:M2}) leverages $\theta \sim P(\cdot \mid T)$ to select actions according to policy $\pi$. Here we make two simplifications of \eq{policy}: first, for our generalization analysis we leave out $\sigma$ so that $\pi$ only depends on $s$, $z$, and $p$. Second, we assume that $\pi$ is a deterministic policy so that $a = \pi(s, z, p)$. We will later show that the empirical results match our theoretical predictions even with these design approximations.

We now have a robot that selects actions according to the latent parameter $\theta$, where $P(\theta \mid T)$ is the learned posterior and $P_0(\theta \mid T)$ is the prior. To complete our analogy with the supervised learning setting we must introduce a loss function that depends on $\theta$. Let cost $C(\theta, f_p) = -R$ be the negative reward of acting based on $\theta$ during a single interaction when the other agent has dynamics $f_p$. Without loss of generality, we again normalize this cost so that $0 \leq C(\theta, f_p) \leq 1$. At this point we have a parallel between RILI and the PAC-Bayes setting: Table~\ref{tab:analogy-PAC} reviews the correspondence. We will therefore apply \eq{pac_bound} to reach our generalization result for RILI robots.

\p{Generalization Result} Consider a RILI robot trained with $N$ partners that have dynamics $\mathcal{S} = \{f_{p_1}, \ldots, f_{p_N}\}$. The robot has learned a Gaussian posterior $P(\theta \mid T)$ that maps sequences of low-level interactions to robot behavior. Let the robot's weights be fixed, so that no new learning occurs. Given the robot's average cost $C_{\mathcal{S}}$ across the seen agents $\mathcal{S}$, for any $\delta \in (0,1)$ the robot's average cost $C_{\mathcal{P}}$ across the entire distribution of latent dynamics $\mathcal{P}(f_p)$ will be less than:
\begin{dmath}\label{eq:gen_bound}
    C_{\mathcal{P}}(P) \leq C_{\mathcal{S}}(P) + \sqrt{\frac{\text{KL}(P \mid\mid P_0) + \log\left(\frac{2\sqrt{N}}{\delta}\right)}{2N}}
\end{dmath}
with probability at least $1 - \delta$. Here $\text{KL}(P \mid\mid P_0)$ is the Kullback-Leibler divergence between the learned posterior over the latent strategy and dynamics $P(\theta \mid T) = \mathcal{N}(\mu, \sigma \mid T)$ and the prior $P_0(\theta \mid T) = \mathcal{N}(0, 1 \mid T)$. We measure the average cost across the known latent dynamics using:
\begin{equation} \label{eq:G1}
    C_{\mathcal{S}}(P) = \frac{1}{N} \sum_{k = 1}^N \frac{1}{|T_k|} \sum_{T_k^i \in T_k} \mathop{\mathbb{E}}_{\theta \sim P(\cdot \mid T_k^i)} C(\theta, f_{p_k})
\end{equation}
Similarly, the average cost across the distribution $\mathcal{P}$ of other agents is:
\begin{equation} \label{eq:G2}
    C_{\mathcal{P}}(P) = \mathop{\mathbb{E}}_{f_p \sim \mathcal{P}} \mathop{\mathbb{E}}_{T \sim P(\cdot \mid f_p)} \mathop{\mathbb{E}}_{\theta \sim P(\cdot \mid T)} C(\theta, f_p)
\end{equation}
Reading Equations~(\ref{eq:G1}) and (\ref{eq:G2}) from left to right, we first take the expectation over the other agent's latent dynamics and then consider the sequences of interactions $T$ that are likely for that specific agent. Finally, we find the expected cost across the latent parameters $\theta$ sampled from the learned posterior $P(\theta \mid T)$. Remember that \eq{G1} is evaluated using measured data from our interactions with the $N$ known agents, while the robot does not evaluate \eq{G2} in practice. Overall, this result enables designers to generalize the performance of their learned RILI system across populations of users. Given that the robot is interacting with a population of humans that have latent dynamics drawn from an underlying distribution $\mathcal{P}$, the robot can leverage its performance with $N$ of those humans to provide a probabilistic lower bound on the expected reward across the entire population.

\begin{algorithm*}[]
\caption{RILI: Robustly Influencing Latent Intent}
\label{alg:rili}
    \begin{algorithmic}[1]
        \State Randomly initialize the robust prediction networks $\mathcal{E}_z, \mathcal{E}_p, \phi, \mathcal{D}$
        \State Randomly initialize the policy and critic networks
        \State Initialize empty replay buffer $\mathcal{B}$
        \State Initialize $z^i \gets \textbf{0}, \sigma_z^i \gets \textbf{0}, p^i \gets \textbf{0}$
        
        \For{interaction $i = 1, 2, \dots$}
                \State Sample a batch of {$m+1$} consecutive experiences {$({T}^m, {\tau}^{m+1}) \sim \mathcal{B}$}
                \State Initialize an empty {batch} of history {${h}^m$}
                \For {{${\tau}$ in ${T}^m$}} 
                    \State Pass {${\tau}$} through $\mathcal{E}_z$ and get the corresponding {batch of} $z$
                    \State Update the history {batch} {${h}^m \gets {h}^m \cup z$}
                    \State {Calculate the loss $\mathcal{L}_z$}
                \EndFor
                
                \State Pass {batch of history} {$h^m$} through $\mathcal{E}_p$ to get $p$
                \State Pass {batches of}  {$z^m$, $\tau^m$, and $p^m$} through $\phi$ to get  {$z^{m+1}$}
                \State {Calculate the loss $\mathcal{L}_\phi$}
                \State {{Calculate total loss $\mathcal{L}$ given by \eq{total_loss}}}
                \State {Calculate the gradient of total loss $\mathcal{L}$}
                \State Update the autoencoder networks $\mathcal{E}_z, \mathcal{E}_p, \phi, \mathcal{D}$
                \State Update the critic and policy networks
            
            \State Collect interaction $\tau^i$ using policy $\pi(a \mid s, z^i, \sigma_z^i, p)$
            \State Update the replay buffer $\mathcal{B} \gets \tau^i$
            \State Get the recent history $T^i$ from $\mathcal{B}$
            \State Estimate recent strategies $z \gets \mathcal{E}_z(\tau)$ for each $\tau \in T^i$
            \State Predict latent dynamics $p \gets \mathcal{E}_p(h^i)$ where $h^i = \{z^i, \ldots z^{i-m+1}\}$
            \State Predict latent strategy $(z^{i+1}, \sigma_z^{i+1}) \gets \phi(\mathcal{E}_z(\tau^i), p, \tau^i)$            
        \EndFor
    \end{algorithmic}
\end{algorithm*}

\section{Implementing RILI}
\label{sec:alg}

Our overall formalism is visualized in \fig{method} and outlined in Algorithm~\ref{alg:rili}. At its heart RILI is built of multiple neural networks: the strategy encoder, the dynamics encoder, a strategy decoder, the predictor, the policy, and the critic. The specific implementation of these models is largely left up to the designer. In this section we provide further details about how we instantiated RILI during our simulations and user studies: we believe that the listed details should serve as a starting point for other designers. Code for our proposed algorithm is available here: \url{https://github.com/VT-Collab/RILI_co-adaptation}

\p{Strategy, Dynamics, and Predictor} The strategy encoder $\mathcal{E}_z$ was a fully-connected network with $2$ hidden layers of size $64$. Recall that $\mathcal{E}_z$ maps low-level observations $\tau$ into a latent strategy $z \in \mathcal{Z}$: in our experiments we used a $10$-dimensional latent strategy space. The strategy decoder $\mathcal{D}$ then takes $z$ and the state-action trajectory $\xi$ and attempts to reconstruct the observed rewards. We similarly implemented $\mathcal{D}$ as a fully-connected network with $2$ hidden layers of size $64$.

The dynamics encoder $\mathcal{E}_p$ inputs a sequence of $m$ consecutive latent strategies $h = \{z^i, \ldots, z^{i-m+1}\}$ and outputs the latent dynamics $p$. For our experiments we set $m=4$, so that the robot attempted to infer the other agent's dynamics from the last four interactions. We designed $\mathcal{E}_p$ as a fully-connected network with $2$ hidden layers and $64$ units per layer. Importantly, $\mathcal{E}_p$ outputs a Gaussian mean $\mu_p$ and standard deviation $\sigma_p$ over the $10$-dimensional latent dynamics space. The predictor $\phi$ was constructed like the dynamics encoder. Model $\phi$ inputs vectors $\tau$, $z$, and $p$ and outputs a Gaussian posterior over the next latent strategy $z^{i+1}$. We programmed $\phi$ as a fully-connected network with $2$ hidden layers of size $64$, and used the reparameterization trick to sample $z^{i+1}$ from a normal distribution with mean $\mu_z$ and standard deviation $\sigma_z$. 

\p{Robot Policy} To perform off-policy model-free reinforcement learning we applied the soft actor-critic (SAC) algorithm \cite{haarnoja2018soft}. The actor (i.e., the policy) and critic were fully-connected networks with $2$ hidden layers of size $256$. We used the $\tanh(\cdot)$ activation throughout the architecture, including the policy and critic networks. To train the representation and reinforcement learning modules we employed two separate Adam optimizers: for robust prediction the learning rate was $1$e$-3$ and for SAC the learning rate was $3$e$-4$.


\section{Simulations} \label{sec:simulations}

In this section we perform controlled experiments to compare our proposed algorithm to state-of-the-art baselines. We leverage three simulated environments established by prior work: within each environment the ego agent interacts with one other agent {across interactions with $H=10$ timesteps}, and the dynamics of this other agent change stochastically between interactions. The resulting simulations have agents who respond to the robot at every interaction, agents who only adapt to some robot behaviors, and agents who stick to their plan and ignore the robot's actions entirely. The ego agent cannot observe the true strategy or dynamics of their partner and must co-adapt over repeated interactions. We compare RILI to a reinforcement learning baseline that does not learn an embedding of the other agent, as well as approaches that combine representation and reinforcement learning while assuming that the dynamics of the other agent are constant.

In our first experiment (Section~\ref{sec:sim1}) the robot learns to influence and coordinate with an unknown distribution of other agents across all three environments. Next, in Section~\ref{sec:sim2} we test the model's capacity to remember old partners: i.e., after learning alongside more than $250$ other agent dynamics, can the robot still coordinate with the original users? In Section~\ref{sec:sim3} we explore the ego agent's ability to adapt to unexpected, out-of-distribution dynamics: i.e., if the robot has interacted with other agents sampled from distribution $\mathcal{P}$, how will it perform with new agents not drawn from $\mathcal{P}$? Finally, in Section~\ref{sec:sim4} we empirically support our probabilistic lower bound on RILI performance. We recognize that these simulations alone do not necessarily capture how each algorithms will perform with actual humans. We therefore use our controlled simulations to complement the user studies in Section~\ref{sec:user_study}. {We also extend these simulations in the Appendix, where we interact with increasingly erratic partners and more complex tasks.}

\p{Baselines} We include four baselines for comparison: 
\begin{itemize}
    \item \textbf{Oracle}. This best-case robot has direct access to the other agent's strategy $z$.
    \item \textbf{SAC} \cite{haarnoja2018soft}. This robot uses only reinforcement learning, and does not learn a representation of the other agent. \textbf{SAC} is equivalent to \textbf{RILI} when the robot's policy is conditioned on state but not $z$, $p$, or $\sigma$.
    \item \textbf{LILI} \cite{xie2020learning}. This related approach learns a latent representation of the other agent's strategy and then conditions the robot's policy on $z$. However, \textbf{LILI} assumes that all other agents follow the same underlying dynamics $z^{i+1} = f(z^i, \tau^i)$. 
    \item \textbf{SILI} \cite{wang2021influencing}. This recent method is an extension of \textbf{LILI} that explicitly encourages the ego agent to stabilize the other agent's latent strategy, i.e., the robot tries to drive $z^{i+1} = z^i$. Like \textbf{LILI}, \textbf{SILI} assumes that all users respond with the same dynamics.
\end{itemize}

\subsection{Simulation Environments}\label{sec:env}

\begin{figure*}
    \centering    \includegraphics[width=2\columnwidth]{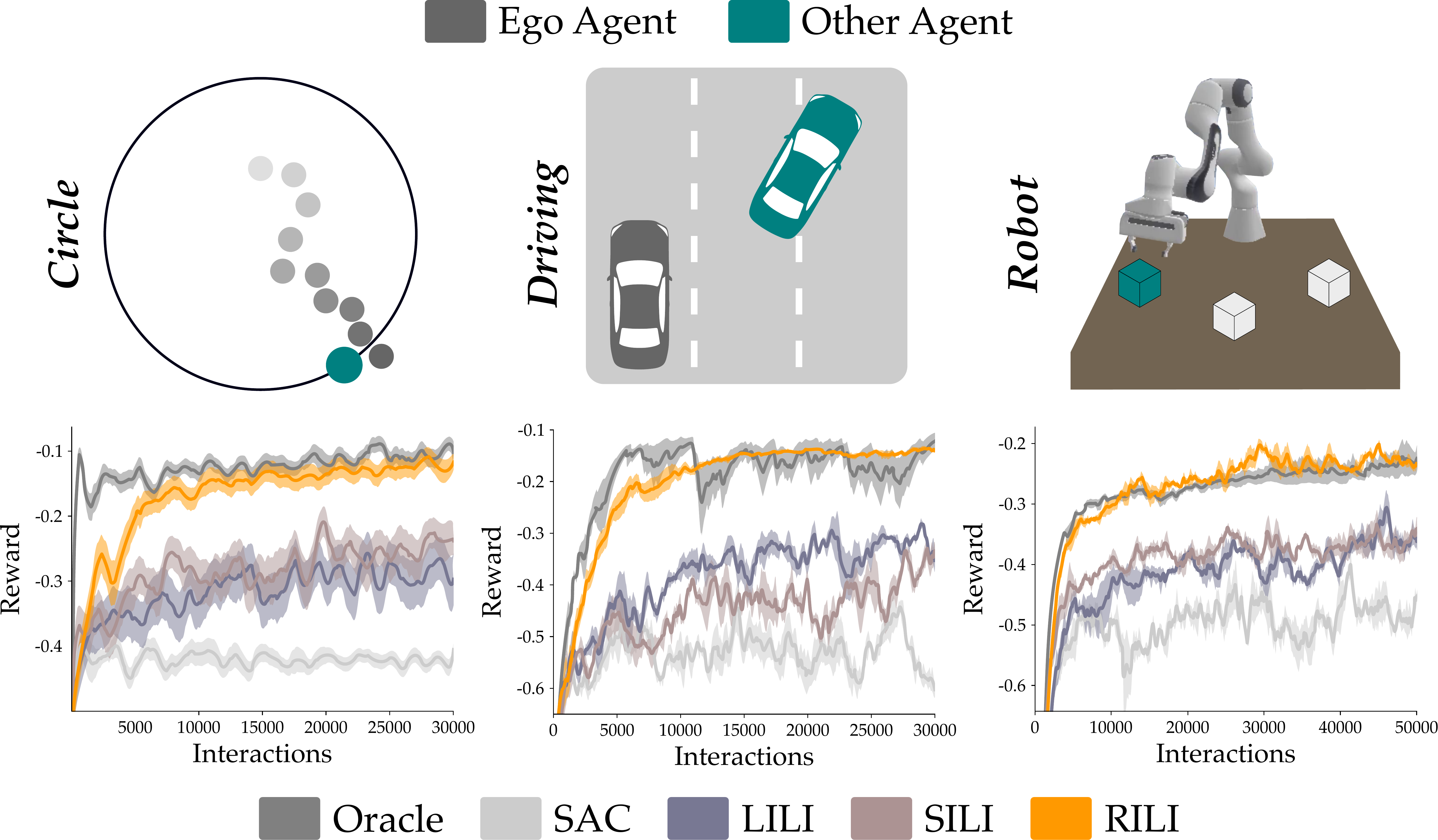}
    \caption{Setup and results from our first experiment (Section~\ref{sec:sim1}). (Top) Environments where the ego agent interacts alongside one other agent. In \textit{Circle} the other agent selects their hidden location, in \textit{Driving} the other agent chooses the lane they merge into, and in \textit{Robot} the other agent picks their desired block. This other agent changes their policy between interactions (e.g., in \textit{Circle} the other agent chooses a new hidden location), and the rules that the other agent uses to make these policy changes also shift stochastically (e.g., in \textit{Circle} the other agent may switch from moving clockwise to moving counter-clockwise). (Bottom) The ego agent's average reward as a function of interaction number. Shaded regions show the standard error across three trials. \textbf{Oracle} always knows the other agent's latent strategy and exhibits best-case performance.}
    \label{fig:sim_1}
\end{figure*}

The experiments in this section were performed on three environments with continuous state-actions spaces: \textit{Circle}, \textit{Driving}, and \textit{Robot} (see~\fig{sim_1}). We selected these environments to remain consistent with the baselines most relevant to our approach \cite{wang2021influencing,xie2020learning}. Each environment consists of an ego agent and another agent: the other agent's policy may change between interactions, and the ego agent does not know either the other agent's policy or their dynamics.

We programmed $N$ different latent dynamics for the other agent in each environment. For Circle, Driving, and Robot we included the dynamics described in \cite{xie2020learning,wang2021influencing}: these dynamics cause the other agent to react to either some or all of the ego agent's behaviors. We then added new dynamics where the other agent ignores the robot and follows a stationary policy. Finally, we created a separate Circle environment where the other agent's dynamics are stochastically sampled from a continuous distribution (we refer to this as \textit{Circle-N}).

\p{Circle} This environment is a pursuit-evasion game \cite{vidal2002probabilistic} with two-dimensional states and actions. During each interaction the ego agent (pursuer) attempts to reach the other agent (evader); however, the ego agent cannot observe the other agent's position. Between interactions the other agent changes their location by moving around the circle. Here {the other agent's latent strategy} $z$ could represent their location and {their latent dynamics} $f_p$ captures how the other agent updates their position between interactions.

For Section~\ref{sec:sim1} we programmed the other agent with $N=4$ possible dynamics. In \textit{Dynamics 1} the other agent moves counter-clockwise when the ego agent lands outside the circle, and otherwise moves clockwise \cite{xie2020learning}; in \textit{Dynamics 2}
the other agent moves clockwise when the ego agent moves outside the circle, and otherwise it does not move \cite{wang2021influencing}. For \textit{Dynamics 3} and \textit{4} the other agent moves counter-clockwise or clockwise regardless of how the ego agent behaves.

\p{Circle-N} This environment is similar to Circle. But instead of designing four latent dynamics, we sample the other agent's dynamics from a continuous distribution $\mathcal{P}$. More specifically, we sample a step size from $-\pi$ to $\pi$ radians. Between each interaction the other agent moves around the circle with the current step size. This environment is especially useful for simulation since we can always sample different step sizes, leading to a potentially infinite number of latent dynamics. 

\p{Driving} In this environment the speeding ego agent is trying to pass a slower driver. The ego agent's state is its $(x,y)$ position, the ego agent's action is its steering angle, and the reward function encourages the robot to minimize steering angle while avoiding a collision. The other agent may change lanes as the ego agent approaches: perhaps a collaborative agent gets out of the robot's way, while a competitive agent merges into the robot's lane. Here {latent strategy} $z$ could represent the lane the other agent merges into and {latent dynamics} $f_p$ captures how the other agent changes lanes in response to the robot's driving.

\begin{figure}[]
    \centering
    \includegraphics[width=1\columnwidth]{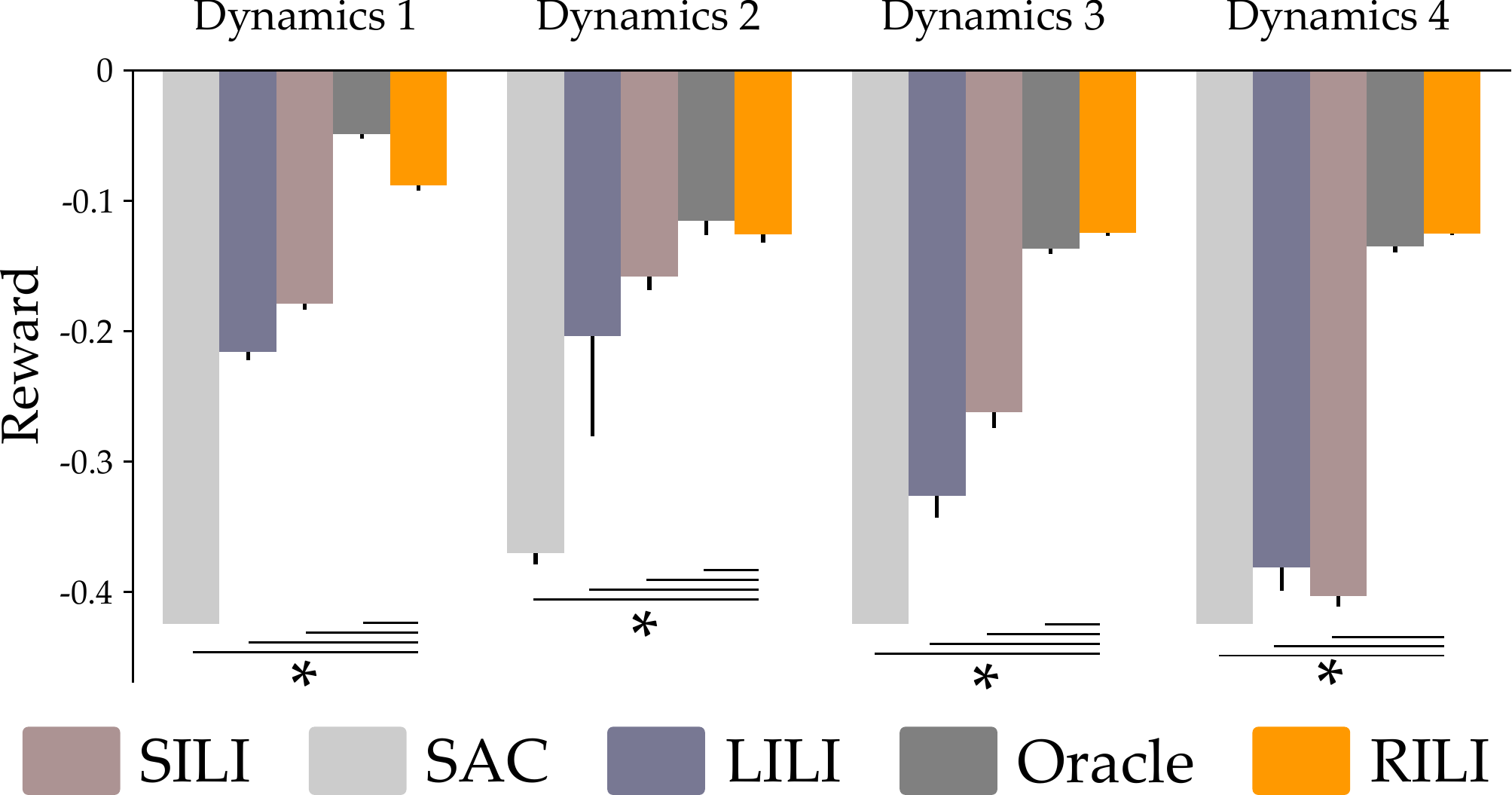}
    \caption{Ego agent's reward across multiple other agents who react to the robot in different ways (Section~\ref{sec:sim1}). For \textit{Circle} in \fig{sim_1} the other agent had four possible dynamics for updating their latent strategy. Here we take the learned models after $30,000$ interactions and roll them out with each of the four dynamics. More negative rewards indicate worse performance, and an $*$ denotes statistical significance $(p < .05)$.}
    \label{fig:sim_1_costs}
\end{figure}

We programmed $N=5$ dynamics for the other agent. In \textit{Dynamics 1} the partner merges into the lane where the ego agent most recently passed \cite{xie2020learning}; in \textit{Dynamics 2} and \textit{3} the other agent moves into the lane that the ego agent occupied earlier in the interaction \cite{wang2021influencing}. Finally, in \textit{Dynamics 4} and \textit{5} the other agent cycles through the lanes either left-to-right or right-to-left regardless of the ego agent's actions.

\p{Robot} In our final environment the ego agent is a simulated Franka Emika robot arm, and the other agent picks one of three goals that it wants the robot to reach. The ego agent does not know which goal the other agent has in mind, and must learn to predict the other agent's choice. The ego agent's state is its end-effector position, actions are end-effector velocities, and the robot's reward is the negative distance between its end-effector and the target object. {Here $z$ could represent the other agent's desired goal and $f_p$ captures how the other agent updates their choice between interactions.} Because of the robot's initial position and the location of the goals, the robot receives higher rewards when the other agent chooses the right-most goal.

We program the other agent with $N=4$ dynamics for choosing targets. In \textit{Dynamics 1} the other agent changes their goal to move away from the robot’s end-effector; in \textit{Dynamics 2} the other agent keeps the same goal if the robot goes to the left of that target, and otherwise moves away from the robot \cite{wang2021influencing}. Finally, in \textit{Dynamics 3} and \textit{4} the other agent cycles clockwise or counter-clockwise through the three goals without responding to the robot.

\begin{figure}[h]
    \centering
    \includegraphics[width=0.8\columnwidth]{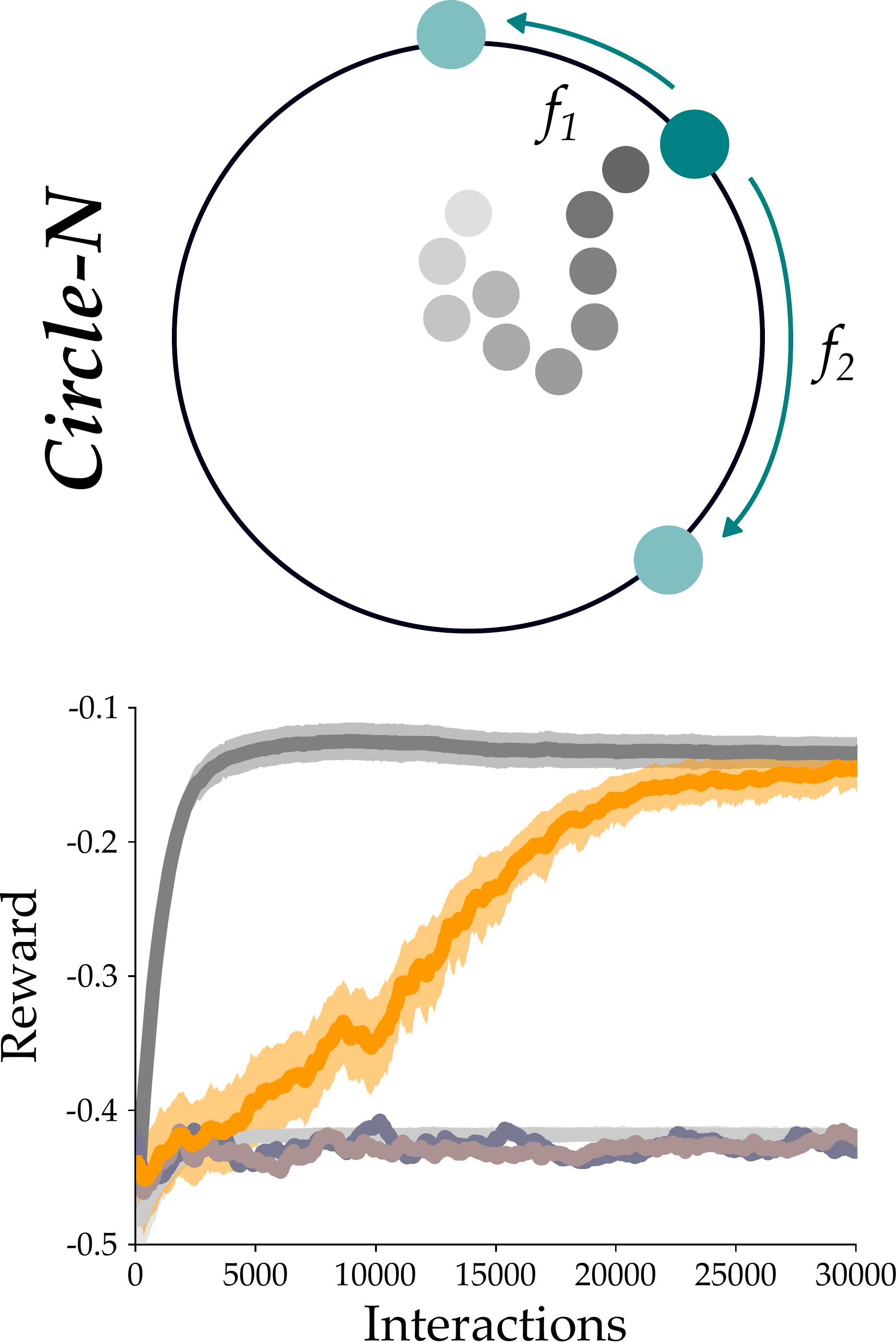}
    \caption{Results from our first experiment in the \textit{Circle-N} environment (Section~\ref{sec:sim1}). (Top) Like \textit{Circle}, the ego agent and the other agent are playing a pursuit-evasion game. The other agent moves their hidden location around the circle between interactions. Here the latent dynamics of the other agent are sampled from a continuous uniform distribution, where each step size $[-\pi, \pi]$ is equally likely. For instance, with dynamics $f_1$ the other agent moves by $+\pi/4$ between interactions, while with $f_2$ the other agent moves $-\pi/2$. (Bottom) The ego agent's reward vs. interaction number. The shaded region is the standard error across three trials.}
    \label{fig:sim_1_dist}
\end{figure}
\begin{figure*}[t]
    \centering
    \includegraphics[width=2\columnwidth]{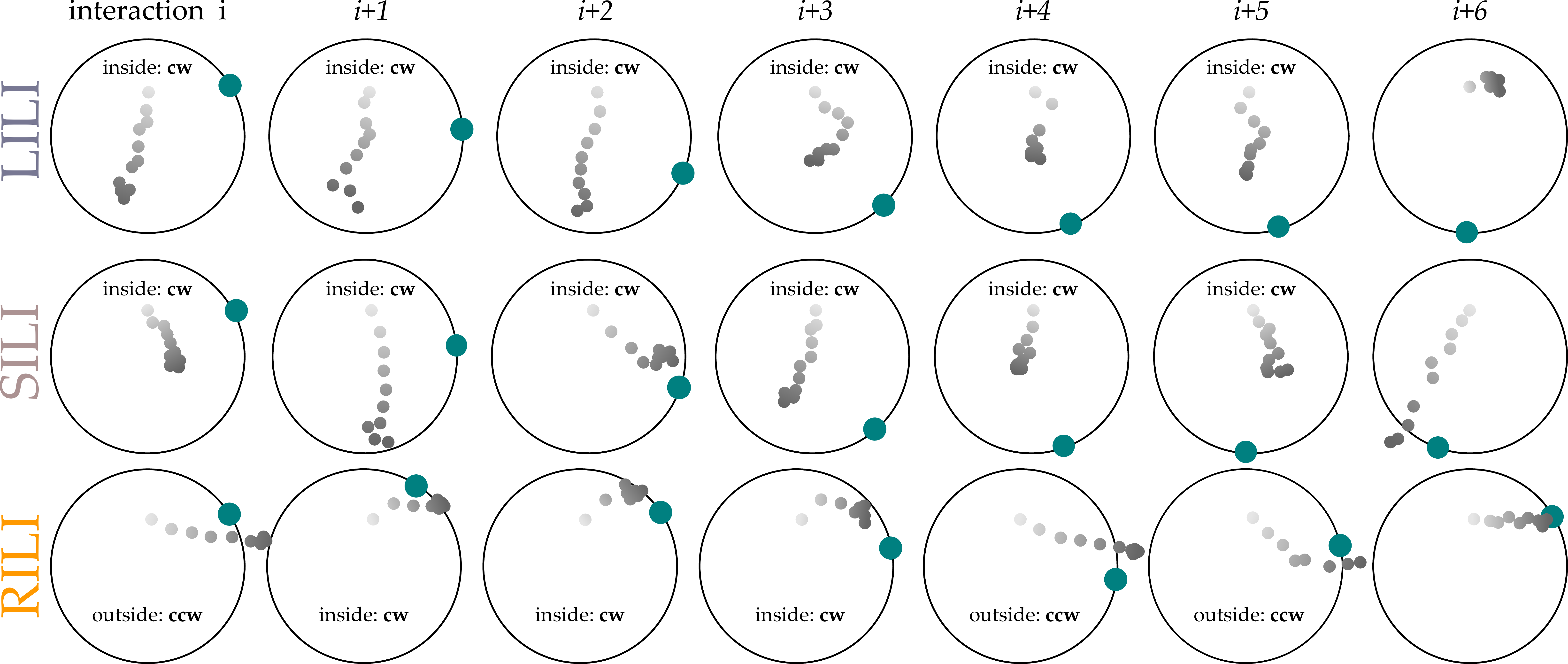}
    \caption{Example of the ego agent and other agent interacting in \textit{Circle} (Simulation~\ref{sec:sim1}). The other agent is using latent dynamics from \cite{xie2020learning}. If the ego agent ends the interaction outside the circle, the other agent moves counter-clockwise; but if the ego agent ends inside the circle, the other agent moves clockwise (\textit{Dynamics 1}). Here the ego agent has previously interacted multiple different latent dynamics, and is now learning alongside an agent using \textit{Dynamics 1}. (Top Row) \textbf{LILI} ends each interaction inside the circle and fails to influence the other agent. (Middle Row) \textbf{SILI} behaves similar to \textbf{LILI}. (Bottom Row) \textbf{RILI} has learned to \textit{influence} the other agent by switching between outside and inside of the circle. Because the ego agent has trapped the other agent close to its start position, \textbf{RILI} is able to reach higher rewards than the alternatives.}
    \label{fig:sim_behavior}
\end{figure*}

\subsection{Coordinating with Changing Agents} \label{sec:sim1}
In our first experiment we pair the ego agent with one other agent whose dynamics change throughout the simulation. The ego agent starts from scratch: this robot has no prior experience, and must learn to successfully complete the task despite the other agent's shifting strategy and dynamics. Across Circle, Driving, and Robot environments the other agent's dynamics change between interactions with a $1\%$ probability. We test \textbf{Oracle}, \textbf{SAC}, \textbf{LILI}, \textbf{SILI}, and \textbf{RILI} three times for the same number of interactions in each environment.

Our results are displayed in \fig{sim_1}. These plots show the ego agent's reward as a function of interaction number. Recall that \textbf{Oracle} has direct access to the other agent's strategy $z^i$ (i.e., in the Circle environment \textbf{Oracle} knows the evader's location). As such, we treat \textbf{Oracle} as the \textit{gold standard} --- in the best case, our learning algorithm should match the performance of \textbf{Oracle}. At the other end of the spectrum is \textbf{SAC}: this approach does not learn a model of the other agent, and seeks behaviors that work well on average. For example, in the Circle environment \textbf{SAC} often moves to the center of the circle (minimizing its expected distance to the evader when the evader's location is unknown). Intuitively, \textbf{SAC} serves as the \textit{worst case} baseline. Across Circle, Driving, and Robot we find that \textbf{RILI} converges to the best-case rewards of \textbf{Oracle}, while \textbf{LILI} and \textbf{SILI} perform similarly to one another and achieve rewards closer to the worst-case \textbf{SAC}. 

For the Circle environment we break these results down by dynamics: see \fig{sim_1_costs}. Remember that during this experiment the other agent in Circle has $N=4$ possible latent dynamics. After completing the interactions shown in \fig{sim_1}, we evaluate the learned models over $1,000$ interactions with \textit{Dynamics 1-4}. We observe that \textbf{RILI} matches the gold standard \textbf{Oracle} for each other agent, while repeated-measures ANOVAs show that \textbf{LILI} and \textbf{SILI} perform significantly worse than \textbf{RILI} across the board ($p<.05$). {Interestingly, \textbf{RILI} slightly outperforms \textbf{Oracle} with \textit{Dynamics 3-4}. One explanation for this could be the interplay of learning alongside the four dynamics. In \textit{Dynamics 3-4} the other agent takes larger steps than in \textit{Dynamics 1-2}. It is possible that \textbf{RILI} learned to move farther clockwise or counterclockwise to trap the other agent in \textit{Dynamics 1-2}, and these larger step sizes translated over to \textit{Dynamics 3-4}. By contrast, \textbf{Oracle} always moves directly towards the known other agent; it therefore learns smaller step sizes in \textit{Dynamics 1-2} and must adapt to larger changes in \textit{Dynamics 3-4}.}

Next we repeat the same experimental procedure in the \textit{Circle-N} environment. So far the ego agent has only had to co-adapt to $N=4$ or $N=5$ dynamics. But within \textit{Circle-N} there is a continuous space of possible dynamics, meaning that each time the ego agent samples from $\mathcal{P}$ it leads to new, previously unseen dynamics. \fig{sim_1_dist} displays how each algorithm performs across three trials. As before, the other agent's dynamics change between each of the $30,000$ interactions with a $1\%$ probability, meaning that the ego agent interacts with roughly $300$ different dynamics. With this increased number of latent dynamics the differences between the algorithms becomes more pronounced: \textbf{RILI} converges to \textbf{Oracle} while \textbf{LILI} and \textbf{SILI} perform on par with \textbf{SAC}. Put another way, \textbf{LILI} and \textbf{SILI} match the performance of a naive robot that always goes to the center of the circle. {We highlight this naive behavior in \fig{sim_behavior} and compare it to the influential behavior displayed by \textbf{RILI}.} Overall, our results suggest that \textbf{RILI} learns to coordinate with agents that shift how they respond to the robot.

\subsection{Revisiting Previously Seen Dynamics}\label{sec:sim2}
If \textbf{RILI} rapidly adapts to coordinate with new agent dynamics, one might ask whether we are also \textit{forgetting} the dynamics that the robot has already seen. Imagine a person who works with the robot one day and then comes back to collaborate a few weeks later: we would hope that the robot both (a) retains the behaviors that led to high rewards with this specific human and (b) transfers other effective behaviors that it has learned from more recent users. We have some indication that the \textbf{RILI} agent can remember and coordinate with multiple dynamics in \fig{sim_1_costs}. But in \fig{sim_1_costs} the robot is continually interacting with the same $N=4$ dynamics throughout the learning process, and there are only four different dynamics to remember. 

\begin{figure}[t]
    \centering
    \includegraphics[width=1\columnwidth]{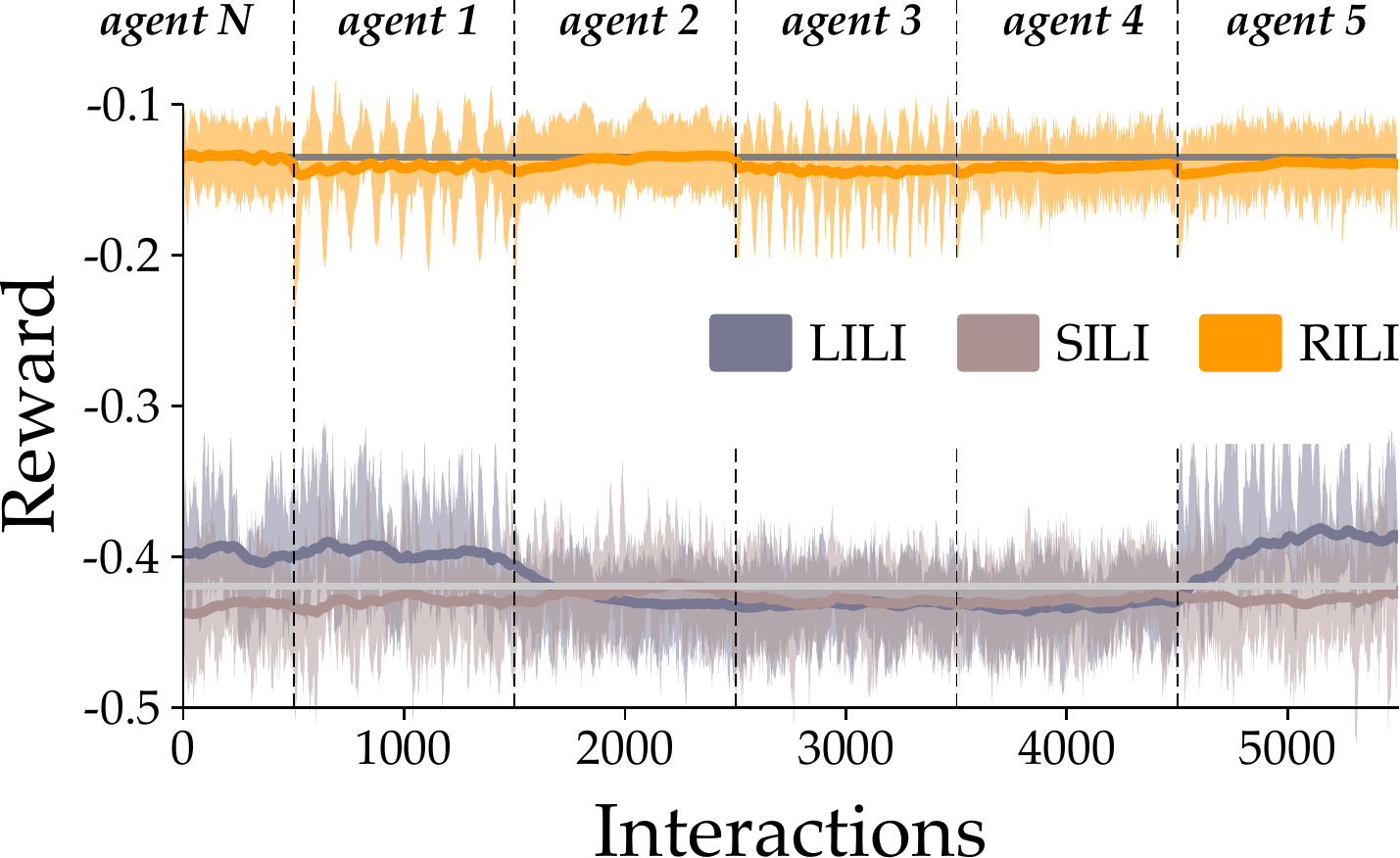}
    \caption{Results from our second experiment (Section~\ref{sec:sim2}). Using the procedure from Section~\ref{sec:sim1} we first train the ego agent with roughly $300$ other agent dynamics drawn from \textit{Circle-N}. We then roll out the final learned models with the first five other agents the robot encountered. The robot interacts with each of these agents for $1000$ interactions: vertical lines indicate when the other agent's dynamics change. Our results suggest that \textbf{RILI} is able to \textit{remember} past dynamics.}
    \label{fig:sim_3}
\end{figure}

In our second experiment we therefore evaluate the robot's ability to coordinate with old dynamics after interacting with a larger number of new, unique agents. First we use the same procedure as Section~\ref{sec:sim1} to train \textbf{RILI}, \textbf{LILI}, and \textbf{SILI} with other agent dynamics drawn from a continuous distribution in \textit{Circle-N}. After $30,000$ interactions the robot has encountered an average of $300$ different partners. We then freeze the networks and \textit{return} to interact with the first five other agents (i.e., the first five simulated humans who used the robot). Our results are shown in \fig{sim_3}. Here the horizontal dashed lines capture the average rewards for \textbf{Oracle} and \textbf{SAC}, and the vertical lines indicate when we switch from one dynamics to another. We find a clear distinction between \textbf{Oracle} and \textbf{RILI} on one extreme and \textbf{SAC}, \textbf{LILI}, and \textbf{SILI} on the other extreme. Our results suggest that \textbf{RILI} is able --- at least to some extent --- to remember previously seen agents. Because the robot does not continue learning during \fig{sim_3}, this also indicates that \textbf{RILI} can predict the strategies and dynamics of multiple different agents.

\subsection{Out-of-Distribution Dynamics} \label{sec:sim3}
\begin{table*}[]
    \centering
    \begin{tabular}{r|p{12cm}}
        \textit{New Dynamics} 1 & If the ego agents moves closer to the other agent at the end of the interaction then the other agent moves counter-clockwise with step $\pi/3$.\\
        \textit{New Dynamics} 2 & Other agent moves to a random point on the quadrant of the circle opposite to where the ego agent ended last interaction. \\
        \textit{New Dynamics} 3 & Other agent moves to the point on circle opposite to the ego agent. \\
        \textit{New Dynamics} 4 & If the ego agent ends on the right side of the circle the other agent goes to $\theta = \pi$. Otherwise the other agent goes to $\theta = 0$.
    \end{tabular}
    \caption{Crowd-sourced dynamics for the pursuit-evasion game (Section~\ref{sec:sim3}). Online participants responded to a survey asking how they would move in \textit{Circle} to avoid the ego agent. We grouped similar responses and reached four new dynamics for the other agent. We then explored how agents trained in \textit{Circle-N} adapted to these new, unexpected other agents (\fig{sim_4}).}
    \label{tab:new_dynamics}
\end{table*}
Our first two experiments suggest that \textbf{RILI} can learn to coordinate with a set of other agents, provided that the dynamics of those other agents are all drawn from the same underlying distribution $\mathcal{P}$. However, it is reasonable to expect that --- at some point during the robot's lifetime --- it will encounter another agent whose dynamics diverge from everything the robot has previously encountered. In other words, our ego agent will run into out-of-distribution other agents. 

\begin{figure}[t]
    \centering
    \includegraphics[width=1\columnwidth]{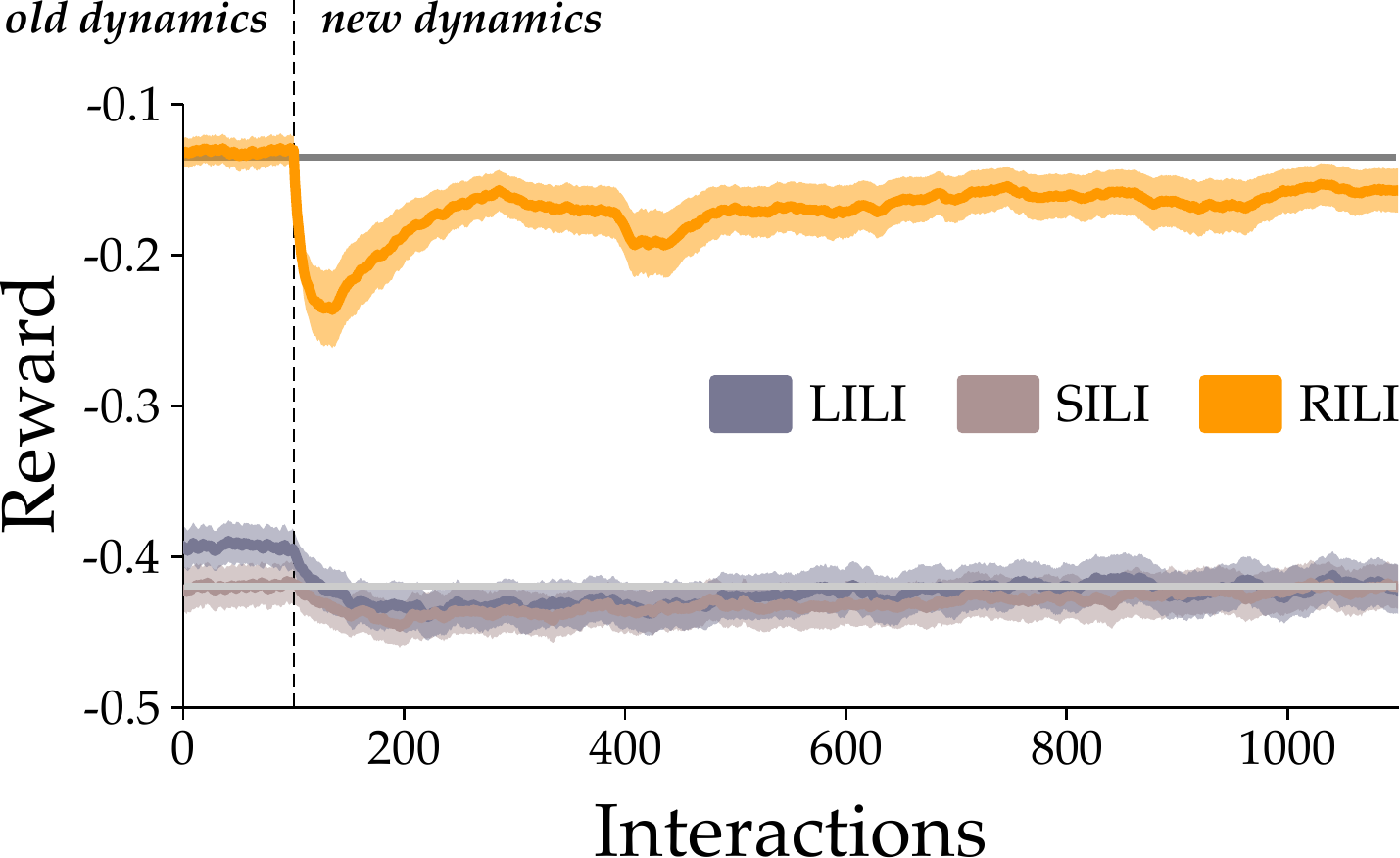}
    \caption{Results for the third experiment (Section~\ref{sec:sim3}). Here we focus on how the ego agent adapts to unexpected other agent dynamics that are \textit{out-of-distribution.} We start with an ego agent trained in \textit{Circle-N} with other agents that move with constant steps of $[-\pi, \pi]$. We then learn alongside four \textit{new dynamics} that are crowd-sourced (see Table~\ref{tab:new_dynamics}). These new dynamics are different from what the robot has seen before; for example, the other agent may react by moving to the point on the circle opposite to where the robot went in the last interaction. We plot the average reward across all four new dynamics over $1000$ interactions. Compared to the baselines, \textbf{RILI} learns to coordinate with these new unexpected dynamics and converges back towards \textbf{Oracle} performance.}
    \label{fig:sim_4}
\end{figure}

In our third experiment we test the ego agent's ability to coordinate with another agent whose dynamics are out-of-distribution. We stick with the \textit{Circle-N} environment, and first trained the robot to coordinate with agents whose step size is sampled from $\pi$ to $-\pi$. We then conducted an online survey to ask human participants for their suggested dynamics. Participants described how they would respond to the ego agent's actions and evade the pursuer: we grouped their responses into $4$ new dynamics. These externally provided, out-of-distribution dynamics are listed in Table~\ref{tab:new_dynamics}. Once collected, we next paired the new other agents with \textbf{RILI}, \textbf{LILI}, and \textbf{SILI} robots. During this experiment the ego agent learned alongside the new dynamics for a total of $1,000$ interactions across three separate trials. Our results are displayed in \fig{sim_4}. Notice that when the new dynamics are introduced the performance of each algorithm drops --- because the other agent is now responding to the robot's behavior in an unexpected way, the ego agent is not immediately sure how to coordinate. But over the course of $1,000$ interactions we again find that the \textbf{RILI} agent converges to the performance of \textbf{Oracle}, and \textbf{LILI} and \textbf{SILI} return to the \textbf{SAC} baseline. Overall, the results from this simulation suggest that learning with \textbf{RILI} not only improves the agent's performance with dynamics sampled from a fixed distribution, but also enables the robot to adapt to new agents who behave in unexpected ways.

\subsection{Empirically Testing Generalization Bounds} \label{sec:sim4}
\begin{figure}
    \centering
    \includegraphics[width=1\columnwidth]{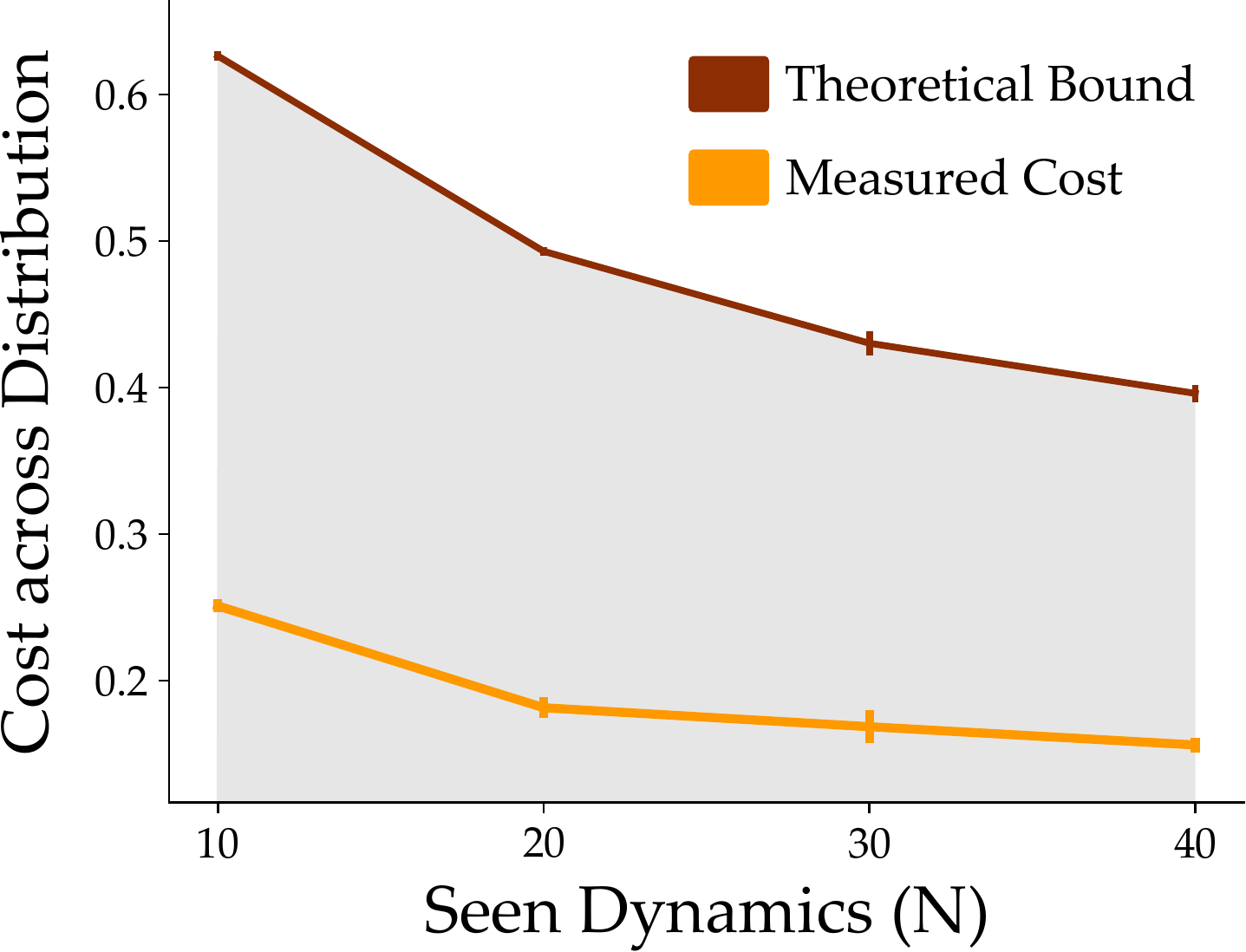}
    \caption{Empirical support for our theoretical bound on \textbf{RILI} performance (Section~\ref{sec:sim4}). In Section~\ref{sec:generalization-bound} we applied PAC-Bayes theory to find an upper bound on cost (i.e., a lower bound on reward). Here we test the resulting bound in simulation. We first trained \textbf{RILI} alongside $N = \{10, 20, 30, 40\}$ dynamics sampled from \textit{Circle-N}. We then used \eq{gen_bound} to plot the probabilistic worst case performance across the entire distribution of dynamics given only $N$ samples. To estimate the true cost we rolled out the learned models with $1,000$ dynamics drawn from distribution $\mathcal{P}$. Our results show that the actual cost (orange) is below the theoretical upper bound (i.e., within the shaded region).}
    \label{fig:sim_2}
\end{figure}

In Section \ref{sec:generalization-bound} we extended PAC-Bayes theory to reach a theoretical lower bound for RILI performance. Given that the ego agent has learned alongside $N$ other agents with dynamics sampled from $\mathcal{P}$, we can leverage \eq{gen_bound} to bound how the ego agent will perform with new agents also sampled from $\mathcal{P}$. In order to reach this result we made two simplifications within \textbf{RILI}: first we assumed that $\pi$ did not depend on $\sigma$, and second we assumed that $\pi$ was deterministic. Here we test the resulting theory while \textit{removing} these assumptions. Put another way, we use our full \textbf{RILI} algorithm with a stochastic policy $\pi(a \mid s, z, p, \sigma)$ and empirically measure whether the algorithm's performance lies within the theoretical bound. For clarity we remind the reader that \eq{gen_bound} provides an upper bound on expected \textit{cost}, which is equivalent to a lower bound on expected \textit{reward}. Here we will measure cost to be consistent with existing PAC-Bayes theory.

Our fourth experiment takes place in the \textit{Circle-N} environment. This experiment has two components: finding the theoretical bound and estimating the actual performance. To obtain the theoretical bound we first train \textbf{RILI} alongside $N=\{10, 20, 30, 40\}$ other agents. For each value of $N$ we perform three runs. We then measure the robot's cost using \eq{G1} and compute the Kullback-Leibler divergence between the model posterior $P(\theta \mid T) = \mathcal{N}(\mu, \sigma \mid T)$ and the prior $P(\theta \mid T) = \mathcal{N}(0, 1 \mid T)$. By plugging these measured terms into \eq{gen_bound} we obtain the theoretical upper bound on cost after working with $10$, $20$, $30$, or $40$ other agents. Remember that this bound indicates how a robot that is trained with $N$ agents will generalize across the entire distribution $\mathcal{P}$. Our second step is to estimate \textbf{RILI}'s actual performance across the entire distribution. To do this we take the \textbf{RILI} model trained after working with $N$ agents, freeze the weights, and roll it out with $1,000$ agents sampled from $\mathcal{P}$. Our assumption here is that by measuring \textbf{RILI}'s performance with these $1,000$ samples we will reach an estimate of the true cost \eq{G2}. We plot both the theoretical upper bound and the estimated true cost in \fig{sim_2}. Our results show that the robot's actual performance is below the upper bound, and as \textbf{RILI} interacts with more agents both the theoretical and empirical costs decrease. This matches our intuition: we expect \textbf{RILI} to perform better across a distribution of dynamics as it interacts with more dynamics from that distribution. Overall, the results from this fourth simulation support the theoretical bound on RILI performance.

\section{User Study} \label{sec:user_study}

Our ultimate goal is a learning algorithm that enables robots to coordinate with and influence \textit{actual} humans. We recognize that humans react to robots, and that different humans adapt to the same robot behaviors in different ways \cite{nikolaidis2017human,goodrich2008human,ikemoto2012physical}. So far our simulations have captured these dynamic interactions in controlled environments (Section~\ref{sec:simulations}). But unlike simulated agents, real humans are noisy and imprecise: their latent strategies and dynamics constantly shift, and a single human may switch between competitive and collaborative interactions. In this section we accordingly perform two user studies with in-person participants recruited from the campus community. In the first study (Section~\ref{sec:user1}) participants play a game of tag with a virtual agent. This agent is trained from \textit{scratch}: the robot starts the experiment without ever having played tag before. Over repeated interactions with multiple humans the virtual robot must learn to anticipate where the human will hide and then reach for that location. In our second study (Section~\ref{sec:user2}) humans build a tower with a Fetch robot arm. Offline we train the robot to build towers alongside simulated users; then during the experiment the robot must learn to adapt to current participant over only $30$ interactions.

\p{Independent Variables} We will compare two robot learning algorithms: state-of-the-art \textbf{LILI} \cite{xie2020learning} and our proposed \textbf{RILI}. Both of these approaches attempt to learn a representation of the other agent and then leverage that representation within the ego agent's policy. But while LILI assumes that all humans react to the robot in the same way (i.e., follow the same underlying rules), RILI learns to robustly predict the other agent's strategy across multiple latent dynamics. We have chosen to use LILI instead of SILI for two reasons: (a) in our pilot studies with simulated humans LILI outperformed SILI and (b) SILI is designed to drive the other agent towards constant strategies, but our user study environments require constantly changing strategies.

\begin{figure*}
    \centering
    \includegraphics[width=2\columnwidth]{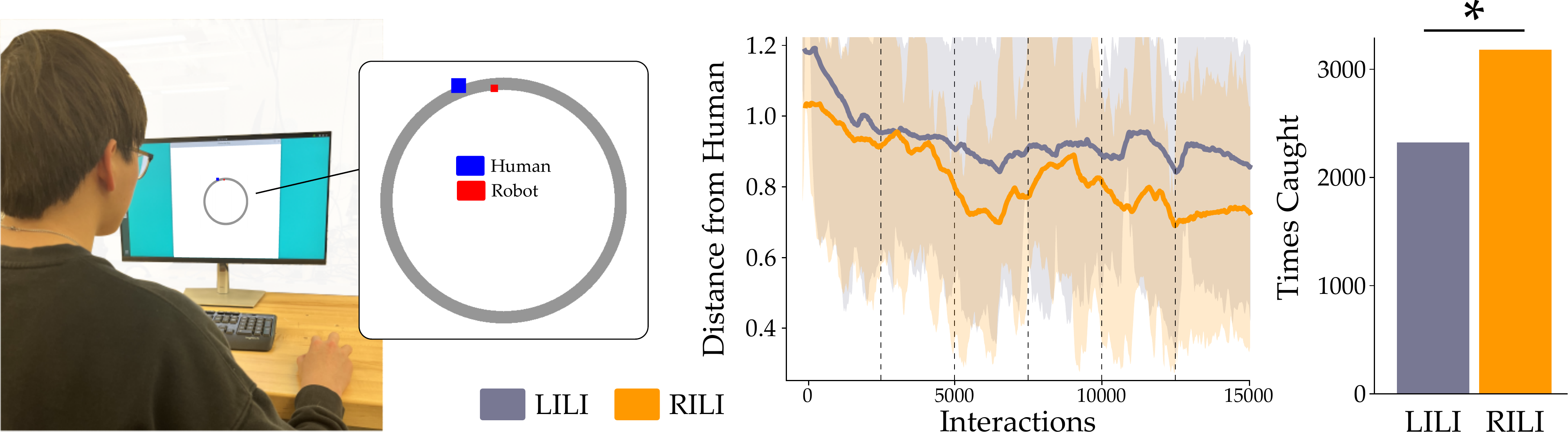}
    \caption{Results from our first user study in Section~\ref{sec:user1}. (Left) Participants visited the lab and played a virtual game of tag. Tag is similar to \textit{Circle} from Section~\ref{sec:simulations}: users moved their cursor around a circle to evade the robot. Each participant was free to choose their dynamics for avoiding the robot, and participants could adapt their dynamics between interactions. (Middle) $22$ participants evaded \textbf{LILI} and \textbf{RILI} agents for a total of $15,000$ interactions. We plot the \textit{Distance} between the robot and human at the end of each interaction. Note that the radius of the circle was one unit, so a \textit{Distance}$>1$ indicates the robot ended more than one radius away from the human. We draw vertical lines every five users, and the shaded regions show standard deviation. (Right) Number of times the robot caught the human. Here a human is caught if the distance between the robot and human is less than $0.5$ at the end of the interaction. \textbf{RILI} caught participants more frequently than \textbf{LILI} ($p < .001$).}
    \label{fig:user_study_1}
\end{figure*}

\subsection{Learning to Play Tag from Scratch} \label{sec:user1}

In our first user study participants interacted with an ego agent in a virtual game of tag (see \fig{user_study_1}). This game is similar to the \textit{Circle} environment from Section~\ref{sec:simulations}. Between each round the human chose where around the circle they wanted to hide, and then the ego agent attempted to predict and reach the human's hidden position. Importantly, the ego agent started from \textit{scratch}: at the beginning of the experiment the virtual robot had no experience playing tag and had not interacted with any simulated users. Participants were not forced to follow any pre-defined rules or patterns of play: we encouraged users to develop their own methods for avoiding the robot, and these dynamics changed both over time and between participants. We measured how \textbf{LILI} and \textbf{RILI} agents were able to coordinate with the changing human participants.

\p{Experimental Setup} Users played a game of tag with a virtual agent by using a clicking interface. The user and virtual agent were point masses in the continuous two-dimensional \textit{Circle} environment outlined in Section~\ref{sec:simulations}. Participants saw a rendering of the environment on a computer screen: the graphical interface displayed their position and the ego agent's {final} position at the end of each interaction. Between interactions users chose where along the circle they wanted to move by clicking on the screen. {Here the user's latent strategy could be where they are hiding on the circle, and their latent dynamics are how they change their hiding location in reaction to the robot. As an example, in one latent dynamics a user might go to locations that maximizes their distance from the ego agent.} The ego agent's state was its $(x,y)$ position, the ego agent's actions were changes in position, and the reward function was the negative Euclidean distance between the ego and the hidden user. The robot maximized its reward by reaching exactly to the human's hidden position. {In each interaction the ego agent started at a position halfway between the center and the top of the circle. The ego agent acted for $H=10$ timesteps to reach the user, and users were shown the ego agent's final position at the end of each interaction.} We emphasize that this environment was continuous, i.e., the human might choose to hide anywhere on the perimeter of the circle.

\p{Dependent Measures} To evaluate how accurately the ego agent (pursuer) caught the human (evader) we measured two things. First, we recorded the distance between the ego agent and the human at the final timestep of each interaction (\textit{Distance}). For reference the radius of the circle was $1$ unit, so a {\textit{Distance} $>1$} indicates that the virtual robot was more than one radius away from the human agent. As a binary measure of success we also recorded the number of times the ego agent caught the user (\textit{Times Caught}). The human was ``caught'' if the distance between the ego agent and the human at the end of the interaction was less than half the circle's radius (e.g., \textit{Distance} $ < 0.5$). The ego agent could \textit{never} observe the human's hidden location, and had to learn to catch this human based on low-level observations of its own states, actions, and rewards.

\p{Participants} We recruited $22$ participants ($4$ female, ages $24 \pm 4.5$ years) from the Virginia Tech community. All participants provided informed written consent following university guidelines (IRB $\#20$-$755$). We conducted a within-subject design. Every participant interacted with both \textbf{LILI} and \textbf{RILI} for the same number of interactions. The order of the methods was counter-balanced between participants, and participants were not told which of the two algorithms learning they were currently interacting with.

\p{Procedure} Before starting the experiment we informed participants about the rules of the game and asked them to play a few practice rounds. Once the experiment started the user was instructed to move around the perimeter of the circle to evade the ego agent. {In our initial trials we found that users always evaded the ego agent by going to the diametrically opposite point. We were concerned that every user might follow this same latent dynamics. To better ensure diversity in user behavior we constrained participants to a step size $\leq \pi / 2$}. Besides this constraint, however, participants were free to choose how they wanted to move in every interaction, i.e., the users could select their own time-varying latent dynamics. As an example, we observed that some users changed their direction of motion frequently to try and throw-off the pursuer, while others moved in a consistent direction away from the robot's last position. We \textit{did not} reset the robot's training between participants --- or even inform the robot that it was now interacting with a new human. Instead, the robot's learning simply resumed once the new participant began to play tag. Overall, we trained each method for a combined total of $15,000$ interactions across all $22$ participants.

\p{Hypothesis} We hypothesized that:
\begin{displayquote}
\textbf{H1.} \textit{Agents that learn from scratch using \textbf{RILI} will more frequently tag adversarial humans than agents that learn from scratch using \textbf{LILI}.}
\end{displayquote}

\begin{figure}
    \centering
    \includegraphics[width=\columnwidth]{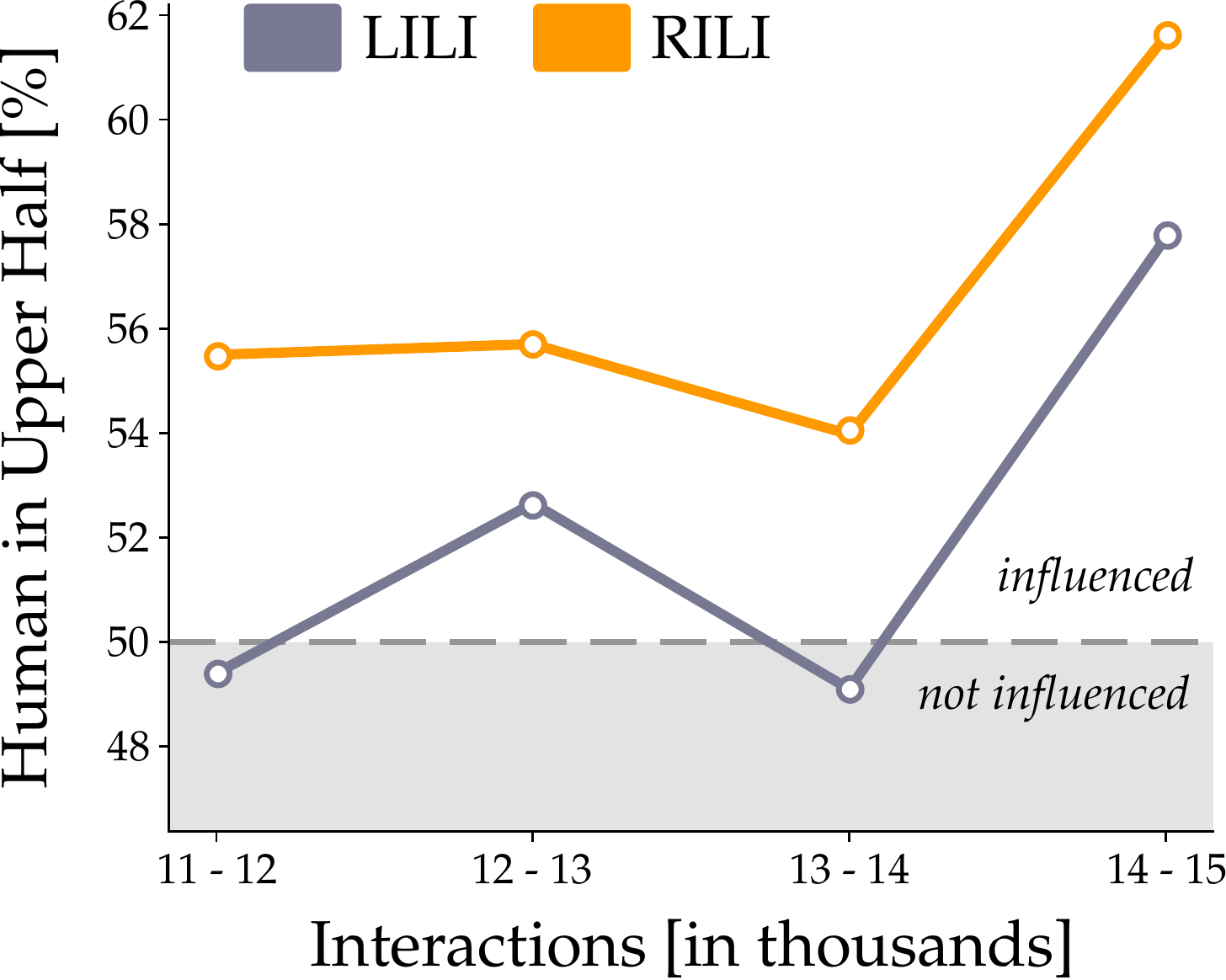}
    \caption{{Comparison of influential behavior in our first user study (Section~\ref{sec:user1}). We plot the percentage of interactions where users hid in the upper half of the circle (see \fig{user_study_1}). Because the robot started along the positive $y$ axis, these human locations are closer to the robot's initial position and lead to higher robot rewards. Hence, the robot should ideally \textit{influence} humans to maximize their frequency of hiding in the upper semicircle. Humans that are not influenced by the robot (i.e., humans that play randomly) will move to the upper semicircle in around $50\%$ of interactions. We find that \textbf{RILI} influenced humans significantly more frequently than \textbf{LILI} over the last $1,000$ interactions ($p < .001$).}}
    \label{fig:influence_user_study_1}
\end{figure}

\p{Results} The results of our first user study are summarized in \fig{user_study_1}. We plot the \textit{Distance} from the robot to the human as a function of interaction number. Remember the ego agent works with each subsequent user without resetting its learning process, so that from the robot's perspective the virtual agent plays $15,000$ interactions with a single other agent. Lower \textit{Distance} scores indicate that the ego agent ended the interaction closer to the human, and as \textit{Distance} approaches zero the robot more accurately predicts and reaches the human's hidden position. We find that \textbf{RILI} reaches the human more accurately than \textbf{LILI} throughout the user study. To determine how many times the robot catches the evading human, we next plot \textit{Times Caught}. This binary metric counts up the number of times the virtual robot ended within a pre-defined distance from the human, and is summed across all $15,000$ interactions. Here \textbf{RILI} catches the human $37\%$ more frequently than \textbf{LILI}. Wilcoxon signed-rank tests reveal that the difference is statistically significant: $Z = -12.819$, $p < .001$. Overall, these results support hypothesis \textbf{H1} and indicate that \textbf{RILI} robots learn to coordinate from scratch more effectively than a state-of-the-art baseline.

\medskip

\noindent{
\textbf{Does RILI Influence Actual Humans?} The overall objective of \textbf{RILI} is to co-adapt to dynamic agents. As part of that objective, in Section~\ref{sec:M2} we designed \textbf{RILI} to learn \textit{influential} policies. We here analyze whether \textbf{RILI} actually results in influential behavior when co-adapting alongside human users. As a reminder, the robot starts every interaction in the $x$-$y$ position $(0, r/2)$, where $r$ is the circle radius, and the robot's reward is the negative of its Euclidean distance from the user's position. Intuitively, if the evading humans are in the \textit{upper half} of the circle the robot can reach them more quickly. Put another way, the robot can maximize its long-term rewards by \textit{influencing} the users to choose hiding locations in the upper half of the circle. We quantify the robot's influence using the metric \textit{Human in Upper Half}, which counts the percentage of interactions where the user hid in the upper semicircle. If the ego agent did not influence users we would expect \textit{Human in Upper Half} to converge around $50\%$. In \fig{influence_user_study_1} we plot the results over the last $4000$ interactions of the user study. From this plot we observe that \textbf{RILI} influenced humans to hide in the upper half more consistently than the \textbf{LILI} baseline. A Wilcoxon signed-rank test revealed that users were influenced significantly more frequently by \textbf{RILI} than by \textbf{LILI} over the last $1000$ interactions ($Z = -6.16, p < .001$). These results suggest the \textbf{RILI} does actually result in robot policies that \textit{influence} humans with changing latent dynamics.}

\subsection{Rapidly Adapting to Build Towers} \label{sec:user2}

\begin{figure*}
    \centering
    \includegraphics[width=2\columnwidth]{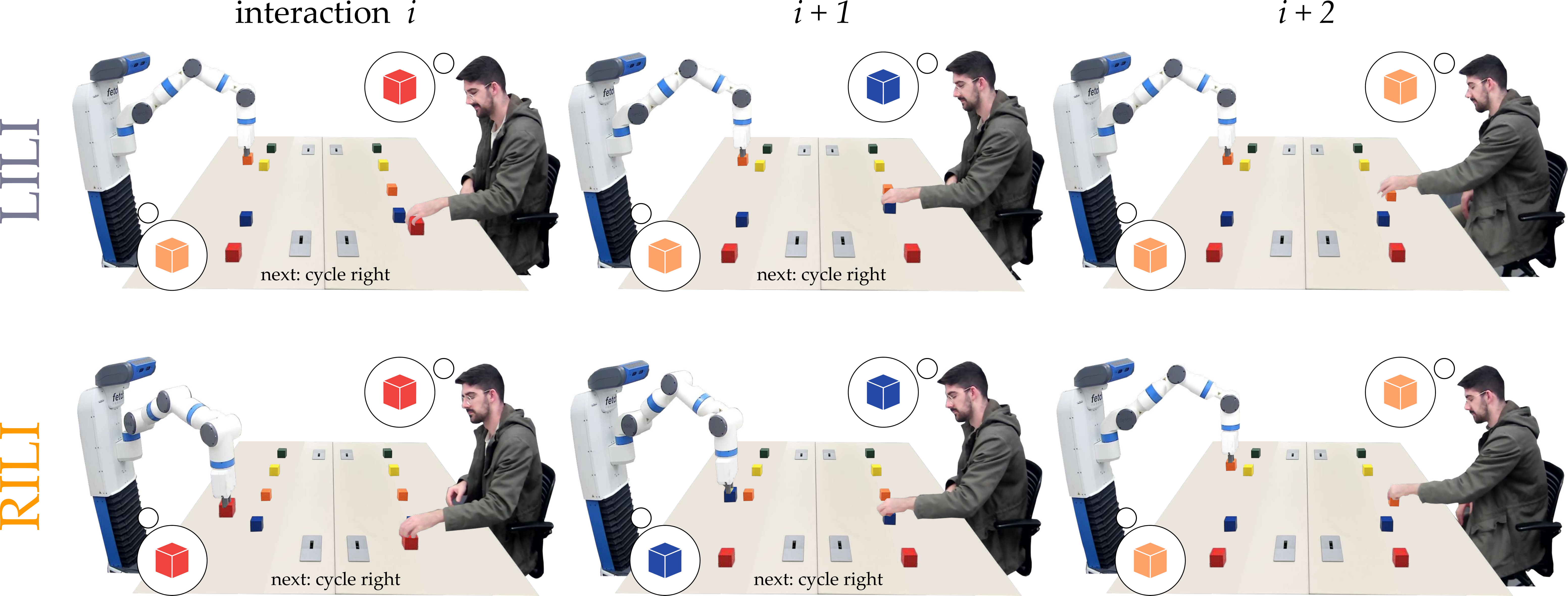}
    \caption{Example interactions between the robot and human during our second user study (Section~\ref{sec:user2}). In each interaction the human and robot pick one block to add to their towers; the robot gets higher rewards for picking the same block as the human. The robot cannot observe the human's choices and only measures its own state, actions, and rewards. This particular user is choosing blocks in order while ignoring the robot's behavior. (Top Row) \textbf{LILI} always reaches for the middle block. (Bottom Row) \textbf{RILI} adapts to the user and matches their block at each interaction.}
    \label{fig:user_study_2_behavior}
\end{figure*}

Our first user study suggests that --- when a robot encounters a new interactive task --- the robot can leverage \textbf{RILI} to gradually adapt to the task and the human's behaviors. But learning both the task and how to coordinate from scratch is challenging: we conducted Section~\ref{sec:user1} in a virtual environment so that we could maximize the number of interactions with humans and collect as many experiences as possible. In more realistic industrial settings we anticipate that manufacturers will \textit{pre-train} the robot using prior knowledge of the task. Accordingly, we here return to our motivating example from \fig{front}: a human and Fetch robot arm work together to build towers. During each interaction the human and robot add one block to their respective towers, and the robot is rewarded for building the same tower as the human. {Here the block the user chooses in a given interaction is captured by their latent strategy, and how they choose their next block is represented by their latent dynamics.} We assume that the robot knows about the task ahead of time. Instead of starting from scratch, we pre-train the robot offline by learning to complete the task with simulated humans. We then put this policy to the test and start to work with actual participants. Overall, we study whether training \textbf{RILI} with simulated humans and then interacting with actual humans leads to \textit{rapid}, \textit{seamless} adaptation.

\p{Experimental Setup} Our experimental setup is shown in Figures~\ref{fig:front}, \ref{fig:front_2}, and \ref{fig:user_study_2_behavior}. Participants sat across the table from a 7-DoF Fetch Mobile Manipulator. On the table were two rows of colored blocks (one row for each agent), and during the $i$-th interaction the human and robot each picked up one block from their row and added it to their tower. {To keep the experiments under $1$ hour in length and maintain subject interest, users did not physically assemble the towers. Instead, the user and robot simply indicated which block they would add to their tower.} The robot's state was its end-effector position, the robot's action was a change in end-effector position, and the robot's reward function was the negative Euclidean distance between the user's chosen block and the robot' end-effector. Put another way, the robot was motivated to pick up the same blocks as the participant. We gave the robot a bonus reward if users picked up the block on their right; because of the configuration of the robot arm, this far right block was the easiest for the robot to grab. {Each interaction lasted $10$ timesteps, and at every timestep the robot would move towards the block it wanted to pick up.}

Before the start of the user study we pre-trained \textbf{LILI} and \textbf{RILI} agents in the tower environment. Instead of interacting with actual users the robots learned alongside a simulated human. We designed three different latent dynamics for this simulated human to leverage: a \textit{competitive} human that picks up the block farthest away from the robot's last choice, a \textit{collaborative} human that picks up the same block the robot chose last time, and an \textit{independent} human who cycles through the blocks while ignoring the robot's behavior. Similar to Section~\ref{sec:sim1} the simulated human's latent dynamics changed with $1\%$ probability between interactions. Both \textbf{LILI} and \textbf{RILI} learned from $10,000$ interactions with these simulated humans. From the robot's perspective the participant's strategy could be the block they select and their dynamics could correspond to how they change their block choice between interactions. 

\p{Dependent Variables} When working with actual participants we recorded the blocks that the human and robot added to their towers at each interaction. Recall that the robot's objective in this study is to build the same tower as the human (e.g., to pick up the same blocks as the human). We therefore measured the robot's performance by counting the number of \textit{Matching Blocks} for each participant. To get a sense of the human and robot policies, we also measured how frequently each of the blocks were chosen (\textit{Block Frequency}). During the experiment the robot never observed which blocks the human picked, and the robot had to learn to coordinate with the human based only on its own states, actions, and rewards.

\p{Participants} We recruited $11$ participants ($11$ male, ages $22.8 \pm 3$ years) from the Virginia Tech community. All participants provided informed written consent prior to the experiment consistent with the university guidelines (IRB $\#20$-$755$). None of the participants in this user study were also participants in the Section~\ref{sec:user1} experiment. We conducted a within-subjects study where each participant interacted with \textbf{RILI} as well as \textbf{LILI}. The order of the methods was counter-balanced across the users (i.e., half the participants started with \textbf{LILI} and the other half started with \textbf{RILI}). Participant completed $30$ interactions with each method for a total of $60$ interactions per participant.

\begin{figure*}
    \centering
    \includegraphics[width=2\columnwidth]{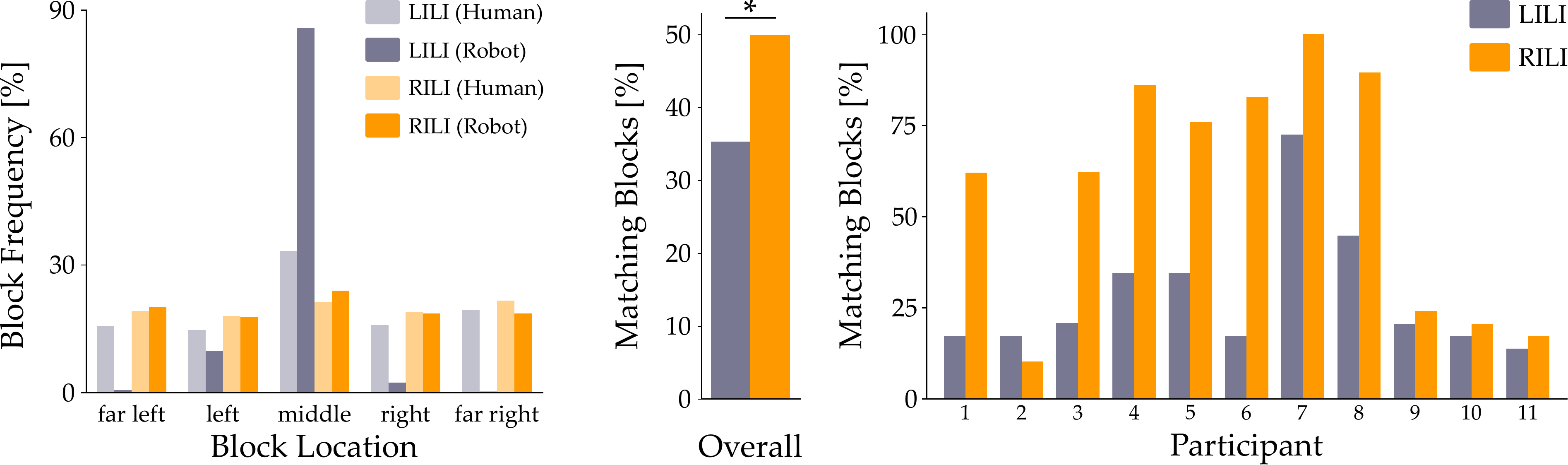}
    \caption{Results from our second user study in Section~\ref{sec:user2}. Here the human and robot are selecting blocks to build towers, and the robot's objective is to match the human's tower. (Left) Frequency the human and robot selected each block. While \textbf{LILI} added the middle block to its tower $86.7\%$ of the time, humans only chose this block in $33.6\%$ of interactions. If the robot is uncertain about the human going for the middle block is a safe play since it minimizes the average distance across all possible human choices. By contrast, we observe that \textbf{RILI} chooses each block at a frequency that roughly matches the actual participants. (Middle) Overall performance of the robot across all $11$ participants. \textbf{RILI} matched the human more often than \textbf{LILI} ($p < .001$). (Right) Breakdown of the robot's performance across each individual participant.}
    \label{fig:robot_choices}
\end{figure*}

\p{Procedure} Before starting the study we explained the setting to users and encouraged them to think about their interaction strategy. {Specifically, the participants were asked to decide on what behavior they will use for selecting blocks; users were instructed to maintain this same pattern throughout the experiment. We explicitly asked  participants \textit{not} to choose blocks at random.} During the experiment the human and robot took turns adding blocks to their respective towers. We placed five different colored blocks in front of the human and the same five different colored blocks in front of the robot (see \fig{user_study_2_behavior}). Within each interaction the human went first and chose their color block; then the robot moved next to selected their own block. Importantly, the robot was never informed which block the human picked. At the end of each interaction the robot observed its rewards and both the robot and the participant reset their blocks for the next interaction. This process was repeated across $30$ interactions per method. Participants were never told how they should choose the blocks and were free to follow their own \textit{personalized} latent dynamics. However, we did instruct the participants to try and maintain consistent dynamics throughout the experiment (i.e., however participants played with the first method they should try to replicate with the second method). 

\p{Hypothesis} We hypothesized that:
\begin{displayquote}
\textbf{H2.} \textit{Robots that use \textbf{RILI} to pre-train with simulated humans and then interact with actual humans will better match the human's towers than \textbf{LILI} robots learning under the same conditions.}
\end{displayquote}

\p{Results} Our results from this final user study are displayed in Figures~\ref{fig:user_study_2_behavior} and \ref{fig:robot_choices}. We also include video from this experiment at: \href{https://youtu.be/WYGO5amDXbQ}{https://youtu.be/WYGO5amDXbQ}

Before summarizing our results we first want to explain the robot behaviors observed during the study. In \fig{user_study_2_behavior} we show  {a sequence of interactions from Participant $4$} with \textbf{LILI} and \textbf{RILI}. This specific participant appeared to choose their blocks in sequential order, and ignored the robot's behavior: in the top row we observe that \textbf{LILI} continually picks the middle block (orange) while on the bottom row \textbf{RILI} has identified the participant's dynamics and matches their choices (red, purple, then orange). {This is supported by the plot for Participant $4$ in \fig{robot_choices}, which shows \textbf{RILI} matches this user's blocks almost twice as often as \textbf{LILI}.} Interestingly, \textbf{LILI} seemed to prefer the middle block across all participants. Looking at \fig{robot_choices}, we notice that the \textbf{LILI} robot selected the middle block $86.7\%$ of the time, and reached for the remaining blocks during only $13.3\%$ of interactions. \textbf{LILI}'s convergence on the middle block is at odds with the human's actual behavior --- looking again at \fig{robot_choices}, when working with \textbf{LILI} the humans selected the middle block only $33.6\%$ of the time and the remaining blocks $66.4\%$ of the time.

So why did \textbf{LILI} incorrectly reach for the middle block so frequently? Remember that the robot's reward function depends on the distance between the robot's end-effector and the human's preferred block. By reaching for the middle block \textbf{LILI} played it safe: going for the middle of the table maximizes the robot's expected reward if --- from the robot's perspective --- the human chooses their next block uniformly at random. Put another way, the middle block makes sense when the robot is uncertain about the human's strategy. By contrast, in Figures~\ref{fig:user_study_2_behavior} and \ref{fig:robot_choices} we observe that \textbf{RILI} reaches for each available block. Here the likelihood of \textbf{RILI} selecting a given block approximately matches the distribution across users: for instance, \textbf{RILI} picked the middle block $24\%$ of interactions while participants working with \textbf{RILI} picked middle in $22\%$ of interactions. Similarly, \textbf{RILI} reached for the far right or left block in $39\%$ of interactions while participants chose these extremes $41\%$ of the time. \textbf{RILI}'s willingness to select other blocks besides the middle indicates that the robot is confident in its prediction of the human's actions --- reaching for blocks on the far right or left only pays off if the robot is guessing correctly.

{With this intuition in mind we now return to our empirical results in \fig{robot_choices}.} We first plot the number of \textit{Matching Blocks} across all $11$ participants. Applying a Wilcoxon signed-rank test, we discover that \textbf{RILI} matched the human's choice significantly more frequently than \textbf{LILI} ($Z = −3.93$, $p < .001$). We next break down these results across the users: for $10$ of the $11$ participants working with \textbf{RILI} resulted in more \textit{Matching Blocks} than working with \textbf{LILI}. {The only anomaly was Participant $2$, who did better with \textbf{LILI} than with \textbf{RILI}. This was likely because Participant~$2$ accidentally used different strategies when interacting with the two methods. When working with \textbf{LILI} this specific user cycled through the blocks (i.e., the participant used latent dynamics that the robot had \textit{seen} during pre-training). In contrast, when interacting with \textbf{RILI} the participant used previously \textit{unseen} dynamics (starting at the ends and moving in). Due to this disparity in the participant’s dynamics \textbf{RILI} had fewer Matching Blocks than \textbf{LILI} for this specific case.} Interestingly, we noticed that \textbf{LILI} performed best with users that \textit{collaborated} with the robot. Participants $7$ and $8$ helped their robotic partner by choosing the same block that the robot picked during the last interaction. Since \textbf{LILI} almost always selected the middle block, these collaborative users converged to also pick the middle block with \textbf{LILI}. But while playing collaboratively did lead to the highest number of matching blocks for \textbf{LILI}, we still found that \textbf{RILI} resulted in even better coordination with these same users.

Overall, the results from our final user study support hypothesis \textbf{H2}. By first pre-training the robot with simulated users, \textbf{RILI} was able to rapidly adapt to actual humans over $30$ interactions. Our comparisons to a state-of-the-art baseline (\textbf{LILI}) suggest that the differences were due to \textbf{RILI}'s ability to anticipate the human's behavior despite the fact that participants followed personalized rules for interaction.

\section{Conclusion}

In this paper we proposed RILI, an algorithmic framework that enables robots to co-adapt alongside non-stationary humans. Learning alongside humans is challenging because (a) humans adapt to robot behaviors, (b) different humans adapt to the same robot behaviors in different ways, and (c) even a single human will inevitably change how they adapt to the robot over time. Put another way, actions the robot has learned to coordinate with one user may fail when that user changes or a new human comes along.

To address these challenges we hypothesized that robots should learn and reason over \textit{high-level representations} of the human. Specifically, we enabled the robot to learn a latent representation of the other agent's policy (i.e., their strategy) as well as a latent representation of how the policy changes (i.e., their dynamics). Our resulting RILI algorithm learns online over repeated interactions using only the robot's low-level states, actions, and rewards. We divided RILI into two parts: \textit{robust prediction}, which learns to anticipate the strategy and dynamics of the current human, and \textit{influential policies}, which harnesses these predictions to intentionally drive the human towards advantageous, co-adaptive behaviors. Given RILI's measured rewards across $N$ humans, we derived probabilistic bounds on RILI's performance with new, previously unseen users. 

To compare RILI to the state-of-the-art we conducted extensive simulations and two user studies. In simulations we compared RILI to representation and reinforcement learning alternatives: our results suggest that RILI is better able to co-adapt alongside agents that constantly change their behavior. We also found that RILI can remember previously seen agents, and rapidly adapt to new agents with unexpected, out-of-distribution dynamics. For our in-person user studies we considered two opposite settings. First, the robot learned to play tag \textit{from scratch} across $15,000$ interactions with adversarial participants. Next, we \textit{pre-trained} the robot to build towers alongside simulated humans, and then rapidly adapted to independent, competitive, and collaborative humans across $30$ interactions.

\p{Limitations} RILI is a first step towards robots that co-adapt alongside non-stationary humans without pre-defined human models or direct observations of the human's behavior. Our simulations and first user study suggest that --- when starting completely from scratch --- RILI will require many human interactions to reach desired performance. This may limit our approach in settings where interactions consume excessive time, materials, or human effort. Our ultimate goal is to co-adapt to the current human as quickly as possible: towards this end, we suggest pre-training the robot alongside simulated agents. These simulated {agents} could emerge from methods such as self-play \cite{carroll2019utility}, population play \cite{jaderberg2019human}, or fictitious co-play \cite{strouse2021collaborating}. Our second user study indicates that --- by training alongside simulated humans --- the robot can rapidly coordinate with actual humans over few interactions (e.g., $<20$ minutes in our study). {We recognize that --- like other learning approaches --- our method's downstream performance is also sensitive to the data available at training time (i.e., the simulated humans used for pre-training). In practice the robot may adapt \textit{quickly} to humans that display latent dynamics similar to the simulated humans seen during pre-training, but the robot may react more \textit{slowly} to humans that display completely new and unexpected behaviors. However, our simulation results in Section \ref{sec:sim4} as well as the user study of learning to play tag from scratch (Section \ref{sec:user1}) suggest that RILI will gradually co-adapt to novel humans.}
\section{Declarations}

\p{Funding} This work was supported by the USDA National Institute of Food and Agriculture, Grant 2022-67021-37868.

\p{Conflict of Interest} The authors declare that they have no conflicts of interest.

\p{Ethical Statement} All user studies were conducted under university guidelines and followed the protocol of Virginia Tech IRB $\#20$-$755$.

\p{Author Contribution} S.P. led the algorithm development, conducted user studies, and wrote the first manuscript draft. D.L. helped develop and implement the method and edited the manuscript.

\p{Acknowledgements}
We thank Joshua Hoegerman for his assistance in running the user studies. 

\bibliographystyle{spmpsci}
\bibliography{citations}

\appendix
\begin{figure*}[h]
    \centering
    \includegraphics[width=2\columnwidth]{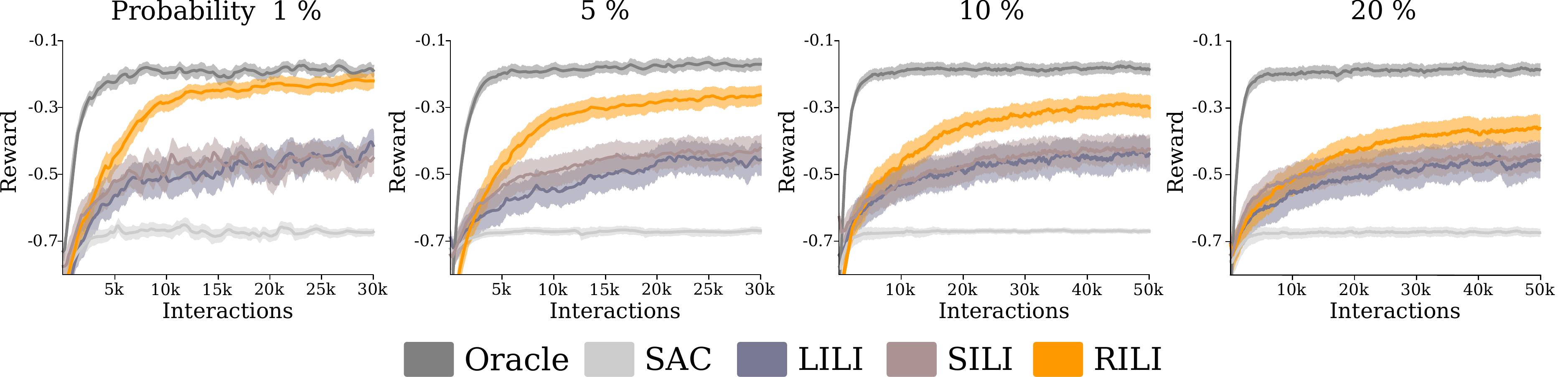}
    \caption{{Interacting with other agents that change their latent dynamics at different frequencies (Appendix \ref{App:rapid_change}). These simulations were performed in the \textit{Circle} environment from Section~\ref{sec:simulations}. (Left) The other agent changes their dynamics with $1\%$ probability after an interaction. This plot corresponds to the \textit{Circle} simulation in \fig{sim_1}. (Middle to Right). The other agent changes their dynamics after $5\%$, $10\%$, or $20\%$ of interactions. Shaded regions show the standard error across three trials.}}
    \label{fig:freq_change}
\end{figure*}

\section{Appendix}

In this appendix we extend the simulation results from Section~\ref{sec:simulations} and \fig{sim_1}. In Section~\ref{App:rapid_change} we vary the rate at which the other partner changes their latent dynamics in the \textit{Circle} environment (i.e., how frequently the other agent switches their response to the ego agent). Next, in Section~\ref{app:complex_env} we scale up the \textit{Robot} environment to include more complex tasks that involve multiple subtasks and an increased number of goals.

\subsection{Coordinating with Rapidly Changing Agents} \label{App:rapid_change}

Recall our simulations from Section~\ref{sec:simulations}. Here the robot learns to coordinate with another agent over repeated interactions, and over time this other agent's behavior can shift as they adapt to the robot. In our experiments we simulated the other agent's adaptation as a probabilistic change in their latent dynamics. More specifically, the other agent's latent dynamics could change between interactions with a $1 \%$ probability. We showed that our method \textbf{RILI} can co-adapt alongside other agents in this setup (see results in Section \ref{sec:sim1}). However, actual humans are much more erratic in their behavior. Can \textbf{RILI} still learn to coordinate if the other agent changes their latent dynamics more rapidly? Here we investigate different rates of adaptation in the \textit{Circle} environment where the robot attempts to reach an evasive other agent. All aspects of the environment are the same as discussed in Section \ref{sec:env} except the probability with which the latent dynamics change. Now we test \textbf{RILI} and the other baselines with another agent that changes with a probability $p = \{1 \%, 5 \%, 10 \%, 20\%\}$. All methods start with no prior experience and are trained for the same number of interactions.

We display our results in \fig{freq_change}. These plots show the robot's reward as a function of the interaction number. \textbf{Oracle} is the baseline that can directly observe the other agent's latent strategy; we therefore treat \textbf{Oracle} as the best-case performance. We see that when the other agent changes less frequently (e.g., $p = 1 \%, 5 \%$) \textbf{RILI} is able to converge close to \textbf{Oracle} performance. As the other agent becomes more erratic (e.g., $p = 20 \%$), we find that the performance of the co-adaptive robot decreases across the board. However, \textbf{RILI} consistently outperforms the baselines, even when faced with these erratic partners. To explain why \textbf{RILI}'s performance decreases as the partner changes more frequently, we note that \textbf{RILI} attempts to predict the other agent's \textit{next} latent strategy based on their history of behaviors. When the other agent rapidly changes their dynamics, it becomes increasingly challenging to anticipate what rules the other agent will follow to select their next strategy.

\begin{figure}[!htb]
    \centering
    \includegraphics[width=\columnwidth]{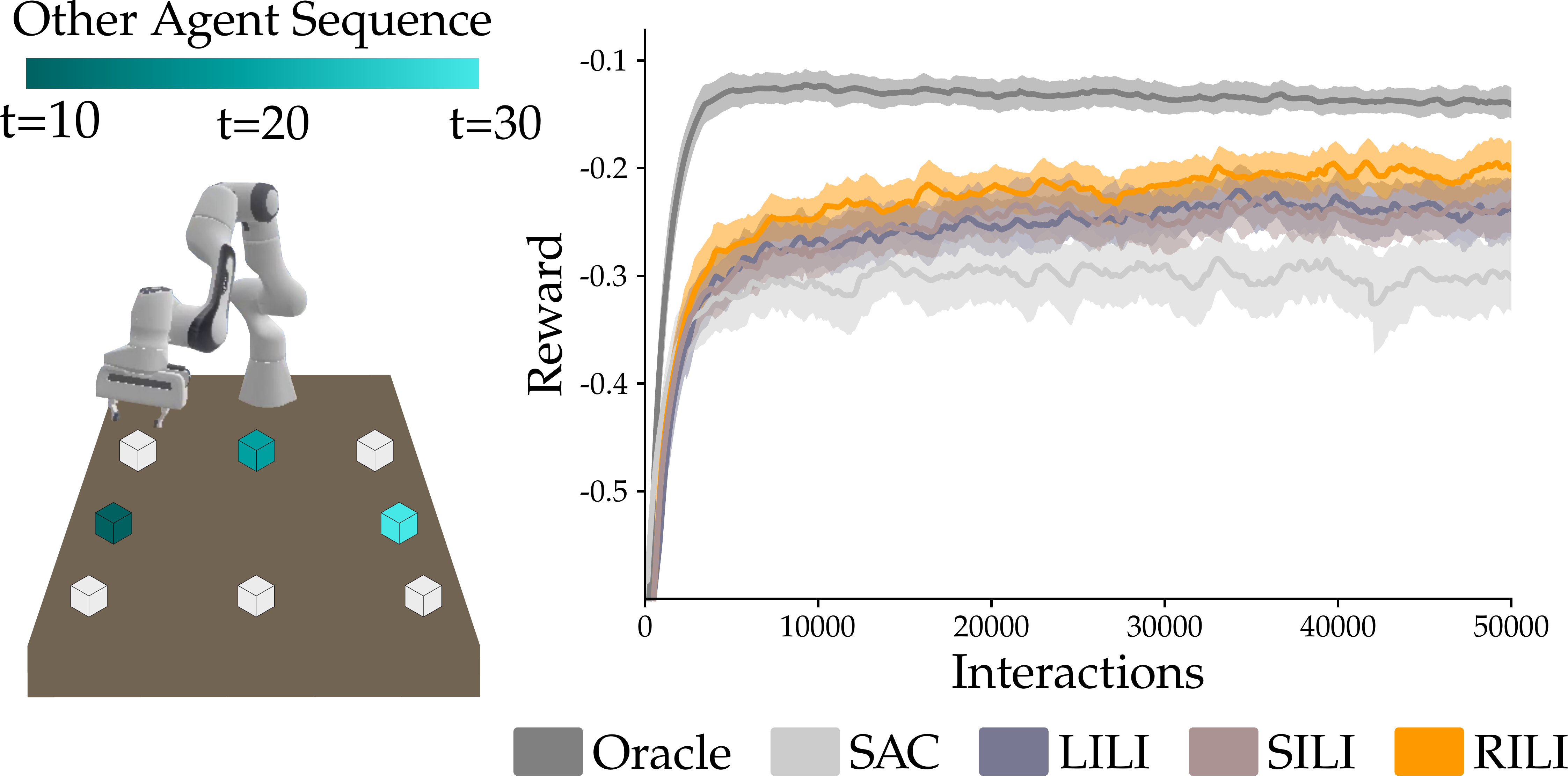}
    \caption{{Increasing the task complexity of the \textit{Robot} environment (Appendix~\ref{app:complex_env}). 
    (Left) Our modified \textit{Robot} environment where the ego agent (robot) interacts with the other agent (human). The other agent selects $3$ desired blocks out of the $8$ possible blocks on the table. The ego agent does not know which blocks the human has selected, and must co-adapt to select these $3$ desired blocks in the same order as the other agent. (Right) Ego agent's average reward vs. the interaction number. The shaded region represents the standard error across three trials.}}
    \label{fig:complex_env}
\end{figure}

\subsection{Coordinating in Environments with Sub-Tasks}\label{app:complex_env}

We have shown that \textbf{RILI} can learn to co-adapt alongside changing agents for multiple tasks --- but what happens as these tasks become increasingly complicated? In this section we scale up the complexity of the \textit{Robot} task from Section~\ref{sec:simulations}. In the new task there are $8$ goals in the robot's workspace. Every interaction the other agent (e.g., the human) selects a sequence of $3$ goals that they want the robot to choose, but the robot cannot observe the other agent's choice or the order in which the human selects these goals. Instead, the robot must learn to anticipate the other agent's choices and the order of selection. We divide the overall interaction into $3$ sub-tasks with $10$ timesteps each (leading to $30$ total timesteps). During every sub-task the robot must pick the correct goal that the other agent selected for that specific sub-task. Thus, the robot's reward function is its distance from the respective goal in the sub-tasks. Depending on their dynamics, the other agent chooses a new sequence of goals at the end of the interaction. We design four different latent dynamics. In every dynamics the other agent chooses three alternate goals each time. In \textit{Dynamics 1} the other agent chooses a new sequence to move away from the robot. In \textit{Dynamics 2} the other agent keeps the same sequence of goals if the robot goes to the left of the third goal in the sequence, otherwise it moves away from the robot. In \textit{Dynamics 3} and \textit{Dynamics 4} the other agent cycles clockwise or counter-clockwise by selecting a new sequence of alternate goals. 
 
We compare \textbf{RILI} with the baselines in \fig{complex_env}. We see that although \textbf{RILI} reaches higher rewards than \textbf{LILI}, \textbf{SILI}, and \textbf{SAC}, it does not converge to the ideal \textbf{Oracle} within $50,000$ interactions. The complexity added by multiple sub-tasks and more intricate latent dynamics makes it challenging for \textbf{RILI} to perfectly model the other agent. However, we emphasize that \textbf{RILI} outperforms the state-of-the-art baselines even as the task complexity scales up.


\end{document}